\documentclass[review,12pt]{elsarticle}

\usepackage{lineno}

\journal{Physics of Fluids}

\usepackage{graphicx}
\usepackage{amsfonts}
\usepackage{amssymb}
\usepackage{float}
\usepackage{latexsym}
\usepackage{amsmath}
\newtheorem{definition}{Definition}
\usepackage{subfigure}
\usepackage[euler]{textgreek}
\usepackage{textcomp}
\usepackage{CJK}

\usepackage{algorithm}
\PassOptionsToPackage{noend}{algpseudocode}
\usepackage[end]{algpseudocode}

\usepackage{booktabs}

\usepackage[table]{xcolor}

\begin{document}

\begin{frontmatter}

\title{Banach neural operator for Navier-Stokes equations}

\author[add1]{Bo Zhang\corref{cor1}}
\ead{bzhang@niu.edu}
\cortext[cor1]{Corresponding author. 
	Address:  
 	DeKalb, IL 60115, USA
 }
\address[add1]{Department of Mechanical Engineering, Northern Illinois University, DeKalb, IL 60115, USA}


\begin{abstract}
Classical neural networks are known for their ability to approximate mappings between finite-dimensional spaces, but they fall short in capturing complex operator dynamics across infinite-dimensional function spaces. 
Neural operators, in contrast, have emerged as powerful tools in scientific machine learning for learning such mappings.
However, standard neural operators typically lack mechanisms for mixing or attending to input information across space and time.
In this work, we introduce the Banach neural operator (BNO)---a novel framework that integrates Koopman operator theory with deep neural networks to predict nonlinear, spatiotemporal dynamics from partial observations.
The BNO approximates a nonlinear operator between Banach spaces by combining spectral linearization (via Koopman theory) with deep feature learning (via convolutional neural networks and nonlinear activations).
This sequence-to-sequence model captures dominant dynamic modes and allows for mesh-independent prediction.
Numerical experiments on the Navier-Stokes equations demonstrate the method's accuracy and generalization capabilities.
In particular, BNO achieves robust zero-shot super-resolution in unsteady flow prediction and consistently outperforms conventional Koopman-based methods and deep learning models.
\end{abstract}


\end{frontmatter}


\section{Introduction}
The inherently dynamic and multidimensional nature of the real world is often elegantly described by partial differential equations (PDEs), which encapsulate the fundamental laws governing the evolution of physical systems over time and space. 
Approximating the solutions to PDEs using either numerical methods or mesh-free neural networks (NNs) is a central task across all corners of computational science and engineering, due to their widespread applications in modeling diverse, complex multiscale and multiphysics systems. 
Despite relentless progress, solving PDEs with traditional numerical solvers is often prohibitively expensive on fine computational meshes. 
On the other hand, classical mesh-dependent deep learning approaches~\cite{Raissi_2019a, Raissi_2020a, Yang_2021a, Karniadakis_2021a, Jin_2021a, Nakamura_2021a, Fukami_2020a, Fukami_2021c, Zhang_2023b, Zhang_2023a} can only learn mappings between finite-dimensional Euclidean spaces, resulting in solutions tied to a fixed discretization~\cite{Li_2020a}. 
As a result, both classical numerical and learning-based methods, albeit of substantive importance, are computationally intractable for solving complex real-world problems. 
Alternatively, learning operator mapping between infinite-dimensional Banach spaces of functions has emerged as a promising paradigm in scientific machine learning (SciML), where deep neural networks (DNNs) are employed as surrogate models to approximate underlying complex PDE systems~\cite{Lu_2021a, Li_2020a, Pickering_2022a, Li_2023a, Lanthaler_2023a, Lyu_2023a, Azizzadenesheli_2024a, Brunton_2024a}. 
Consequently, developing operator regression strategies with robust generalization and extrapolation capabilities, particularly for nonlinear operators and functionals, has become a task of paramount importance in SciML. 

The frontier of operator regression is advancing relentlessly, with neural operators demonstrating remarkable success in transferring solutions across meshes with different discretizations.  
Unlike traditional finite-dimensional models that require mesh-dependent formulations, neural operators are built specifically and predominantly for generalizing across varying input resolutions and parametric dependencies.  
Among a rich diversity of neural operators, deep operator network (DeepONet)~\cite{Lu_2021a} and Fourier neural operator (FNO)~\cite{Li_2020a} stand out as two powerful supervised, data-driven deep learning architectures for efficiently solving PDEs within the emerging field of SciML. 
DeepONet relies on the universal operator approximation theorem to learn various explicit and implicit nonlinear operators with small generalization error. 
In contrast, FNO parameterizes the integral kernel directly in Fourier space, allowing it to efficiently map parametric inputs to PDE solutions with high accuracy and resolution-invariant performance. 
Building upon these foundational architectures, numerous extensions have emerged to broaden their applicability and improve performance under practical constraints.
In order to mitigate the impact of data sparsity in numerous realistic scenarios, physics-informed DeepONet~\cite{Wang_2021a, Goswami_2023a} has been developed.
This exploits the fact that DeepONet outputs are differentiable with respect to input coordinates, enabling the use of automatic differentiation to enforce PDE constraints via regularization.      
For efficiently characterizing extreme events in society and nature, a scalable, model-agnostic, Bayesian-inspired framework based on deep neural operator (DNO)~\cite{Pickering_2022a} was formulated.     
By including two independent DeepONets coupled by residual learning and input augmentation, a multifidelity DeepONet~\cite{Lu_2022a} was proposed to learn the phonon Boltzmann transport equation (BTE), demonstrating a fast solver for the inverse design of BTE problems via combining a trained multifidility DeepONet with genetic algorithm. 
Wen \textit{et al.} presented U-FNO~\cite{Wen_2022a}, which appends a U-Net path in each U-Fourier layer to enrich the representation power of the architecture in higher frequencies information. 
Li \textit{et al.} proposed an implicit U-Net enhanced FNO (IU-FNO)~\cite{Li_2023b}, which incorporates the U-Net into the implicit U-Fourier layer to efficiently predict long-term large-scale dynamics of turbulence. 
Additionally, Lyu \textit{et al.}~\cite{Lyu_2023a} developed a multi-fidelity learning method based on FNO. 

While the aforementioned methodologies have showcased the potential of DeepONet and FNO in approximating complex physical processes, a wide family of new frameworks for operator learning have been explored by recognizing the limitations and computational bottlenecks inherent in DeepONet, FNO and their variants. 
To address FNO's limitations in capturing transient responses and non-periodic signals, Cao \textit{et al.}~\cite{Cao_2024a} proposed the Laplace neural operator (LNO).
This architecture incorporates the pole-residue relationship between input-output spaces in the Laplace domain.
For learning the infinite-dimentional mapping on irregular grids, the neural operator was instantiated via graph kernel network~\cite{Li_2020b}, employing message passing on graph networks.  
For learning non-local operators for dynamics with long-distance relations, Zappala~\cite{Zappala_2024a} \textit{et al.} formulated the neural integral equation (NIE) and the attentional neural integral equation (ANIE), which are learned integral operators that allow to continuously learn dynamics with arbitrary time resolution. 
To reduce the prohibitively high computational costs of many-query evaluations of PDE solutions on multiple geometries, Yin \textit{et al.}~\cite{Yin_2024a} proposed the diffeomorphic mapping operator learning (DIMON) framework. 
DIMON provides a scalable method for learning the geometry-dependent solution operators of PDEs.  
In addition, Xiong \textit{et al.} developed the Koopman neural operator (KNO)~\cite{Xiong_2024a}, which is a mesh-independent neural-network-based solver of PDEs. 
KNO demonstrates enhanced capacity to learn the long-term behaviors of PDEs and dynamic systems. 

Whereas the FNO and LNO learn integral operators via levaraging the Fourier transform (FT) and Laplace transform (LT) to approximate non-local mappings in function spaces respectively, the LT can be seen as a natural extention of the FT since it generalizes the idea of decomposing a time-domain function into a set of sinusoidal basis functions~\cite{Cao_2024a}, while offering a broader framework for dealing with growth, decay and transient behaviors. 
The FT captures periodic structures and global dependencies, while the LT captures exponential growth or decay behaviors. 
In a similar vein, dynamic mode decomposition (DMD)~\cite{Schmid_2010a, Schmid_2022a, Baddoo_2023a} can be understood as a combination of proper orthogonal decomposition (POD)~\cite{Berkooz_1993a, Schmidt_2020a} and FT, and hence the resulting spatial modes are associated with a given temporal frequency accompanied by a growth or decay rate.
Moreover, DMD approximates the linear, infinite-dimensional Koopman operator, which depicts the temporal evolution of observables in a dynamical system and linearizes nonlinear dynamics in an extended function space. 
On the other hand, the future state of the system $u(t)$ can be formulated in a spectral expansion form as a linear superposition of the truncated $r$ DMD modes,
\begin{align}
	u(t) &\approx \sum^{r}_{j=1} b_j \phi_j \exp(\omega_j t) \, ,
	\label{eq:dmd}\,
\end{align}
where $b_j$ is the initial amplitude of the $j$th DMD mode, $\phi_j$ represents the corresponding spatial DMD mode, and $\omega_j = \ln (\lambda_j)/\Delta t$ denotes the continuous time eigenvalue with $\Delta t$ the time step of the snapshots. 
Here, $\lambda_j = \sigma_j + i\omega_j'$ denotes the discrete DMD eigenvalue with $\sigma_j$ dictating growth or decay and $\omega_j'$ determining oscillations. 
Additionally, DMD like the LT captures modes that evolve as $e^{\lambda t}$ and as a result, it can deal with both exponential growth or decay and oscillations. 

Regardless of the theoretical and computational advancements of the FT and LT, both methods face critical limitations when applied to complex real-world systems. 
Specifically, they are ill-suited for noisy and sparse datasets, lack the capacity to effectively handle high-dimensional spatiotemporal data, and cannot extract dominant coherent structures that evolve over space and time. 
Moreover, these classical transforms are fundamentally linear and thus inadequate for capturing nonlinear dynamics inherent in many physical systems. 
They also suffer from computational inefficiency when scaled to large datasets. 
Additionally, the FT is limited to signal decomposition and does not provide a predictive model for forecasting future states, while the LT maps functions to a transformed domain without offering a direct mechanism for temporal prediction.  
Such limitations render efforts of utilizing them to analyze nonlinear, high-dimensional systems moot. 
In contrast, DMD offers several advantages.
It is computationally efficient for large, high-dimensional systems by capitalizing on singular value decomposition (SVD)-based algorithms. 
Furthermore, DMD is able to extract both spatial and temporal modes, making it particularly well-suited for spatiotemporal systems such as turbulent flows, even without prior knowledge of the explicit underlying equations. 
Building upon the strengths of DMD and convolutional neural network (CNN)~\cite{Lecun_2015a, Silver_2016a, Goodfellow_2016a}, a neural operator framework is proposed, defined in infinite-dimensional Banach space and designed to deliver excellent super-resolution and prognostication capacities.  

Early frameworks such as DeepONet~\cite{Lu_2021a} introduced powerful universal approximators for learning nonlinear operators between infinite-dimensional function spaces and achieved impressive performance in various parametric PDEs.
However, DeepONet's branch-trunk architecture is primarily designed for static mappings, offering limited scalability to high-resolution spatiotemporal domains and lacking mechanisms for temporal modeling or interpretable modal analysis. 
Spectral approaches like the Fourier neural operator (FNO)~\cite{Li_2020a} and Laplace neural operator (LNO)~\cite{Cao_2024a} addressed these limitations by learning the kernel of a non-local operator in Fourier or Laplace space. 
FNO parameterizes the solution operator in the Fourier domain, making it effective for globally smooth, periodic problems but less suited for transient or non-perioridc phenomena. 
LNO replaces the Fourier basis with Laplace modes to better represent exponential decay or growth, yet both methods remain difficult to interpret physically and lack explicit temporal modeling. 
Extensions such as U-FNO~\cite{Wen_2022a} and IU-FNO~\cite{Li_2023b} embed U-Net architectures within FNO to enhance multiscale resolution and high-frequency representation. 
These enhancements improve predictive performance but remain fundamentally tied to Fourier-based global convolutions and do not provide interpretable dynamics or explicit sequence modeling. 
The Koopman neural operator (KNO)~\cite{Xiong_2024a} embeds system evolution in a latent space inspired by Koopman theory, enabling long-term forecasting.
However, KNO does not evolve observables directly via Koopman eigenvalues and modes, limiting its interpretability and physical transparency. 
Likewise, neural integral equation (NIE) and attentional neural integral equation (ANIE)~\cite{Zappala_2024a} learn continuous-time kernels over history, and diffeomorphic mapping operator learning (DIMON)~\cite{Yin_2024a} handles geometry-dependent PDE operators via diffeomorphic mappings. 
While these methods address different challenges, none incorporate a physically grounded modal decomposition for dynamic reconstruction. 
In contrast, the proposed Banach neural operator (BNO) explicitly incorporates Koopman spectral theory and DMD into a trainable deep operator learning pipeline. 
BNO evolves dynamics in a Koopman modal space spanned by approximate eigenfunctions, enabling interpretable sequence-to-sequence (seq2seq) forecasting, zero-shot super-resolution, and robust generalization across spatial scales. 
Its CNN-based projection captures complex spatiotemporal dependencies, resulting in a physically grounded, mesh-independent framework for learning operators in nonlinear PDEs such as the Navier-Stokes equations. 
To the best of our knowledge, BNO is the first Koopman-based operator learning framework that systematically unifies spectral reconstruction with end-to-end trainable deep neural architectures for nonlinear PDEs. 

Additionally, wavelet-based methods have been used to solve the Navier-Stokes equations with remarkable accuracy, as demonstrated by recent work employing the Euler wavelet collocation method to simulate staionary flows in complex geometries~\cite{Mohammad_2024a}.
This technique is particularly effective at handling fractional differential equations with exceptional accuracy.
Mohammad and Trounev~\cite{Mohammad_2025a} further investigated the effects of fractal geometries on viscous flow using finite element simulations, revealing that fractal roughness induces distinct vortex shedding patterns and chaotic structures at high Reynolds numbers. 
To capture transitions from laminar to turbulent regimes---along with memory and nonlocal effects---they proposed modeling the dynamics with fractional time derivatives.
While these approaches focus on numerical simulation of complex geometries, our work aims to learn neural operators in a data-driven framework that is capable of generalizing across spatial resolutions and capturing long-term dynamics.
Specifically, the proposed Banach Neural Operator (BNO) integrates Koopman theory and convolutional neural networks (CNNs) to model both global linear structures and local nonlinear interactions, offering a flexible operator-learning framework that can be adapted to complex domains such as fractal boundaries.
Unlike traditional solvers, BNO does not require an explicit meshing or fractional discretization, making it well-suited for fast inference and autoregressive rollout in highly nonlinear flow regimes.
Recent advances in physics-informed neural networks (PINNs) have also shown promise in solving the Navier-Stokes equations by embedding physical laws directly into the training objective~\cite{Botarelli_2025a}.
This approach circumvents traditional computational fluid dynamics (CFD) bottlenecks by avoiding mesh generation and iterative solvers, enabling faster solutions in complex geometries and real-world applications.
In contrast, the BNO does not rely on an explicit formulation of the governing PDEs.
Instead, it approximates the solution operator between Banach spaces purely from spatiotemporal data, facilitating generalization across discretizations and boundary conditions.

Drawing motivation from FNO, BNO is inspired directly by recognizing the predictive reconstruction capability of Koopman theory.  
The main contribution of this article is the proposal of an advanced, expressive deep learning framework---BNO---that learns mappings between infinite-dimensional function spaces in a seq2seq manner, achieving superior generalizabilities irrespective of discretization constraints or mesh dependencies. 
BNO outperforms pure data-driven methods markedly and pushes the boundaries of classical neural operators.   
The most significant are summarized below: 
\begin{itemize}
        \item A novel framework is introduced for operator regression in the Koopman domain, capable of learning complex multiscale and multiphysics dynamics. This paves the way for robust operator learning in SciML. 
	\item BNO is designed with an iterative structure, which can be readily formulated as a recurrent neural network with parameters shared across all layers in the sequence. 
	\item BNO is able to produce high-resolution outputs even when trained exclusively on low-resolution data, enabling zero-shot super-resolution across discretizations.
	\item BNO adopts an attention-inspired deep neural architecture, which accounts for the mixing of information among the input sequence to learn long-range dependencies and capture contextual information beyond local pixel neighborhoods. 
	\item BNO learns dominant spatiotemporal patterns from ranining data and accurately forecasts unsteady flows in a seq2seq manner.  
	\item Dynamic Mode Decomposition (DMD), which approximates finite-dimensional representations of the Koopman operator, is seamlessly integrated into the neural architecture. This enables fully differentiable, end-to-end seq2seq training. 
\end{itemize}

\section{Methods} 
\label{sec:method}
The strength of the proposed Banach Neural Operator (BNO) is demonstrated in approximating highly nonlinear operators arising in the Navier-Stokes equations.
The model integrates a Koopman-based linear approximation with convolutional neural networks (CNNs) and nonlinear activation functions, effectively capturing both global structure and local interactions in the data.

\subsection{Problem setup: operator learning in Banach spaces}
Let $\Gamma \subset \mathbb{R}^d$ be a bounded spatial domain and let $[0,T] \subset \mathbb{R}$ denote a fixed temporal interval.
Define the spatiotemporal domain as $Q := \Gamma  \times [0,T] \subset \mathbb{R}^{d+1}$.
Let $\mathcal{B} = \mathcal{B}(Q;\mathbb{R}^{d_b})$ and $\mathcal{W} = \mathcal{W}(Q;\mathbb{R}^{d_w})$ be separable Banach spaces of functions defined on Q, where $\mathcal{B}$ and $\mathcal{W}$ can be taken respectively as input and output spaces taking values in $\mathbb{R}^{d_b}$ and $\mathbb{R}^{d_w}$ in operator learning. 
Our overarching goal of neural operator learning is to approximate a nonlinear operator 
\begin{align}
        \mathcal{F}: \mathcal{B} \rightarrow \mathcal{W},  
	\label{eq:map1}\,
\end{align}
which maps an input sequence $b(x,t) \in \mathcal{B}$ to an output sequence $w(x,t) \in \mathcal{W}$, typically representing a spatotemporal solution to a system of PDEs such as the Navier-Stokes equations.
The neural operator approximation is then defined as a parametric nonlinear mapping 
\begin{align}
	\mathcal{G}: \mathcal{B} \times \Theta \rightarrow \mathcal{W} \,,
	\label{eq:map}\,
\end{align}
where $\theta \in \Theta \subset \mathbb{R}^q$ are tunable network parameters that are learned by backpropagating gradients via automatic differentiation. 
The training dataset consists of input-output pairs $\{(b_{i},w_{i})\}_{i=1}^{N}$ where each $b_{i} \in \mathcal{B}$ is a spatiotemporal sequence discretized over a grid $Q_i := \{(x_j,t_k)\}_{j,k} \subset Q$ and $w_i = \mathcal{F}(b_i)$.
The observed input and output sequences are sampled as
\begin{align}
	b_i |_{Q_i} \in \mathbb{R}^{N_x \times N_t \times d_b}, \ \ \ w_i |_{Q_i} \in \mathbb{R}^{N_x \times N_t \times d_w} \,,
	\label{eq:map}\,
\end{align}
where $N_x$ and $N_t$ are the number of spatial and temporal discretization points, respectively.
Let $\mu$ be a propability measure over $\mathcal{B}$, representing the distribution of input sequences.
The operator learning objective is to minimize the expected loss functional
\begin{align}
	\mathcal{J}[\mathcal{F}, \mathcal{G}] &:= \mathbb{E}_{b\sim\mu}[\mathcal{L}(\mathcal{F}(b), \mathcal{G}(b,\theta))] \, 
	\label{eq:functional}\,
\end{align}
where $\mathcal{L}: \mathcal{W} \times \mathcal{W} \rightarrow \mathbb{R}$ is a suitable loss function (\textit{e.g.}, a squared $L^2$ norm over $Q$) that quantifies the error between the ground truth and the predicted outputs.
The optimal neural operator is obtained by solving
\begin{align}
	\theta^{*} & = \arg\min_{\theta \in \Theta} \mathbb{E}_{b\sim\mu}[\mathcal{L}(\mathcal{F}(b), \mathcal{G}(b,\theta))] \,,
 	 \label{eq:optimize}\,
\end{align}
yileding the learned mapping $\mathcal{G}^{*}(\cdot) := \mathcal{G}(\cdot,\theta^{*})$.

Most conventional numerical solvers and state-of-the-art deep learning methods suffer from the discretization-invariant problem, namely, such approaches are mesh-dependent and highly sensitive to the choice of grid discretization.
This limits their practical applications to problems where solutions to complex PDE systems with fine discretization are required. In contrast, the proposed Banach Neural Operator (BNO) addresses this issue by approximating nonlinear infinite-dimensional operators directly between Banach spaces.
By leveraging the Koopman operator framework and an attention-inspired deep neural architecture, BNO learns representations that are inherently resolution-agnostic.
Consequently, BNO is capable of producing solutions $w(x,t)$ at arbitrary spatiotemporal locations $(x,t) \in Q$, including points $(x,t) \notin Q_i$ not seen during training---enabling robust zero-shot super-resolution across different discretizations.

\subsection{BNO architecture and recursive update rule}

\begin{figure}[tbp]
\begin{center}
\includegraphics [width=1\columnwidth]{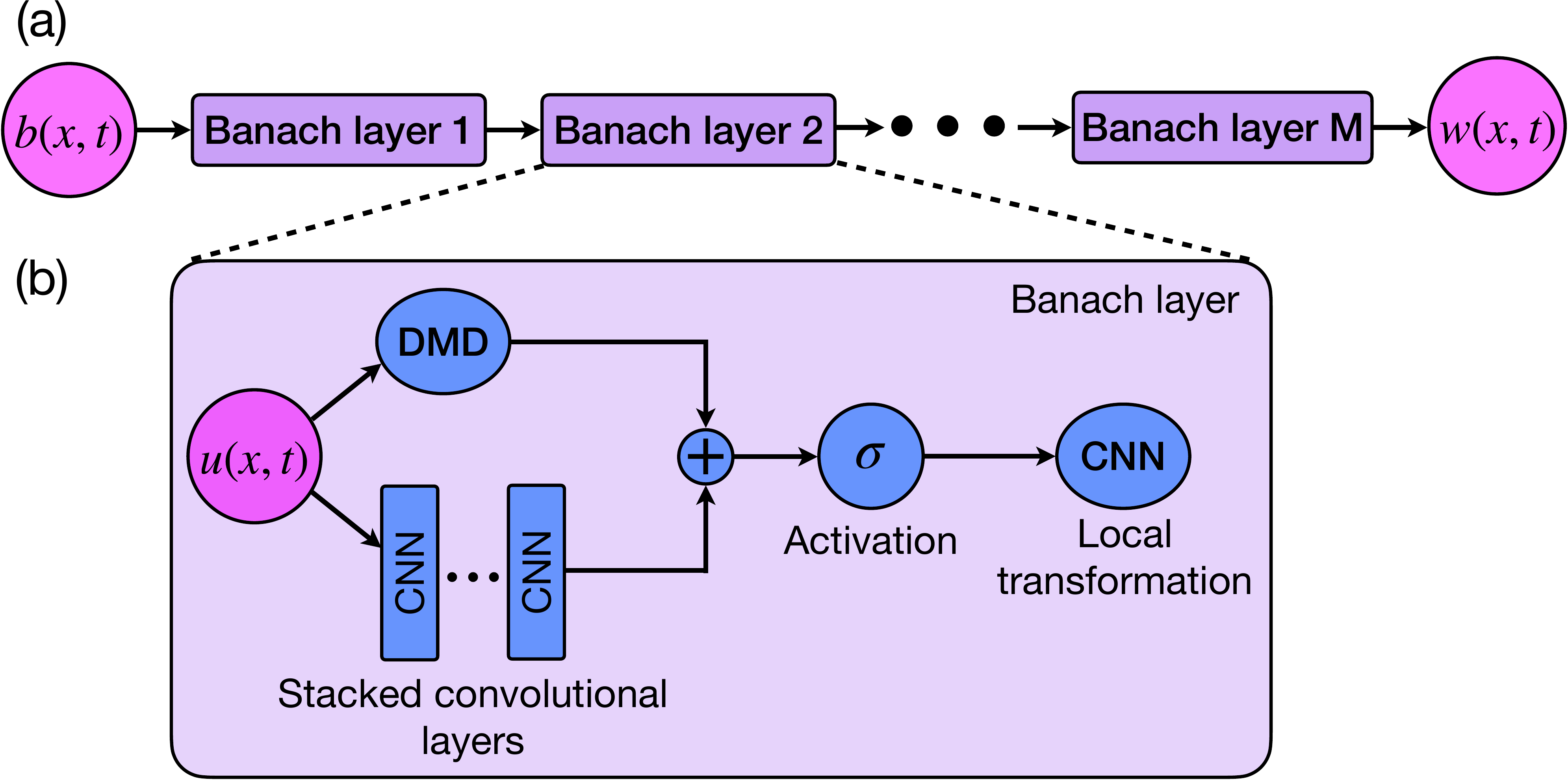}
\end{center}
\caption{Illustration of the BNO architecture. (a) A schematic showcasing the overall architecture of the BNO. We start with an input function $b(x,t)$. 1. Maintain the original dimensionality for the input layer; 2. Apply several layers of Banach neural operators, each composed of Koopman discrete operators, stacked convolutional operators, nonlinear activation functions and local transformations; 3. Finally output $w(x,t)$. (b) Banach layers: start from input $u(x,t)$. Top pannel: Perform DMD analysis to compute the truncated DMD modes and eigenvalues, then reconstruct the future state. Bottom panel: Apply stacked convolutional operators to the input sequence.}
\label{fig:architecture}
\end{figure}  

Similar to FNO, BNO is structured as a sequence of iterative updates 
\begin{align}
         u_0 \mapsto u_1 \mapsto \cdots \mapsto \cdots \mapsto u_M \,,
 	 \label{eq:updates}\,
\end{align}
where each intermediate $u_i: Q \rightarrow \mathbb{R}^{d_u}$ is a spatiotemporal function defined on the domain $Q := \Gamma  \times [0,T]$.
Each $u_i \in \mathcal{W}(Q;\mathbb{R}^{d_u})$ denotes a latent representation of the solution evolving over $M$ Banach layers~\cite{Li_2020a}.
As illustrated in Fig.~\ref{fig:architecture}, the input $b \in \mathcal{B}$ is directly fed into the neural network, maintaining its original dimensionality $u_0(x,t) = b(x,t) \in \mathcal{B}(Q;\mathbb{R}^{d_b})$ for the input layer rather than being lifted to a higher dimensional representation by a neural network. 
Then, the representation $u_i$ is updated iteratively across Banach layers.
Within each layer, the intermediate output is projected back to the target channel dimension via a local transformation $P$, which is parameterized by a CNN with a linear activation function. 
Ultimately, the final output is given by $w(x,t) = u_M(x,t)$. 
For every iteration, the update $u_i \mapsto u_{i+1}$ is accomplished through a non-local Koopman discrete operator $\mathcal{K}$, a projection operator consisting of a series of convolutional layers with a nonlinear activation function in each layer, a local, nonlinear activation function and a local transformation $P$. 
The following definition is then introduced: 
\begin{definition} [Recursive representation update]
The recursive update rule for the representation $u_i \mapsto u_{i+1}$ is defined as
\begin{equation}\label{def:con}
u_{i+1}(x,t) := P  \circ \sigma\Bigg(\bigl(\mathcal{K} u_i\bigr)(x,t) + \sum_{t'=1}^{T}\bigl(\mathcal{C}_{t'}(c_{t'};\zeta_{t'})u_{i}\bigr)(x,t) \Bigg), \quad \forall (x,t) \in Q,
\end{equation}
where $\mathcal{K} \in \mathcal{L}(\mathcal{W}(Q;\mathbb{R}^{d_u}), \mathcal{W}(Q;\mathbb{R}^{d_u}))$ is a bounded linear operator acting on the Banach space $\mathcal{W}(Q;\mathbb{R}^{d_u})$, interpreted as a Koopman operator in this context, $\mathcal{C}_{t'}$ is the CNN operator in the $t'$-th layer with its own learned kernel $c_{t'}(x-y;\zeta_{t'})$ parameterized by $\zeta_{t'} \in \Theta_{\mathcal{C}_{t'}}$, the summation represents a learnable corrector composed of $T$ convolutional layers, $\sigma$ is a nonlinear activation function applied component-wise and $P$ is a local transformation, typically parameterized by a CNN with a linear activation function.
\end{definition}
This recursive update allows the model to propagate spatiotemporal features through the network while simultaneously projecting the evolving representation into a structured feature space. 
The operator $\mathcal{K}$, interpreted as a Koopman operator, acts as a bounded linear transformation on the Banach space $\mathcal{W}(Q;\mathbb{R}^{d_u})$. 
Even though $\mathcal{K}$ is linear, the full neural operator is highly nonlinear since a composition with activation function introduces nonlinearity, analogous to standard neural networks and on the other hand, stacking multiple nonlocal and local operators renders the overall architecture deep, resulting in a deeply composed operator that captures complex dynamics. 
Additionally, the corrector term---represented by a sequence of CNN operators $C_1, \cdots, C_T$ with intermediate nonlinearities---further enhances expressiveness.
Specifically, the corrector is defined by a nested composition of the form
\begin{align}
        u_T = \sigma \circ \mathcal{C}_T \circ \sigma \circ \mathcal{C}_{T-1} \circ \sigma \circ \cdots \circ  \mathcal{C}_{2} \circ \sigma \circ \mathcal{C}_{1}u_i \,,
 	 \label{eq:corrector}\,
\end{align}
which introduces rich nonlinear transformations and allows local refinement of the representation produced by $\mathcal{K}$.

Collectively, these components enable the BNO to generalize convontional PDE solvers by learning operator-valued mappings from data, bypassing the need for mesh-dependent discretizations.

\subsection{Banach neural operator}
Koopman theory provides a global, spectral perspective on dynamical systems. 
In neural operator frameworks, the Koopman operator generalizes convolution operators to complex, infinite-dimensional spaces, enabling nonlocal interactions. 

\subsubsection{Koopman discrete operator via DMD}
DMD is a method that approximates the action of the Koopman operator on a finite-dimensional subspace spanned by a set of observables. 
In this work, the Koopman operator $\mathcal{K}$ in Equation~\ref{def:con} is replaced by the Koopman discrete operator.
\begin{definition} [Koopman discrete operator $\mathcal{K}$]
Define the Koopman discrete operator as 
\begin{equation}\label{def:kdo}
\bigl(\mathcal{K}u_i\bigr)(x,t) = (\boldsymbol{\Phi} \exp(\boldsymbol{\Omega}t) (\boldsymbol{\Phi}^{\dag} u_i))(x), \qquad \forall (x,t) \in \Gamma \times \mathbb{R}^{+}
\end{equation}
where $\boldsymbol{\Phi} \in  \mathbb{C}^{N \times r}$ denotes the matrix with the DMD modes given by its column vectors assuming the domain $\Gamma$ is discretized with $N \in \mathbb{N}$ points, 
$\boldsymbol{\Omega} = \text{diag}(\omega) \in \mathbb{C}^{r \times r}$ is a diagonal matrix with the entries $\omega = \ln (\lambda)/\Delta t$, and $\boldsymbol{\Phi}^{\dag} \in  \mathbb{C}^{r \times N}$ represents the Moore-Penrose pseudo-inverse of $\boldsymbol{\Phi}$. 
\end{definition}
This formulation bridges operator learning with DMD, showing that the temporal reconstruction can be viewed as an application of a Koopman discrete operator that propagates initial conditions forward in time. 
Moreover, the Koopman discrete operator acts as a linear propagator, evolving the observables via a discrete operator, which is explicitly constructed from DMD modes and eigenvalues, making it inherently data-driven and adaptable to complex dynamical systems. 
In essence, the Koopman discrete operator provides a practical approximation of the Koopman operator by restricting it to a finite set of dominant eigenfunctions and eigenvalues identified from data. 

\begin{table}[htbp]
\setlength{\tabcolsep}{.2em}
\caption{Details of the BNO network architecture for a single Banach layer. 
The input is processed in parallel by a CNN-based branch and a Koopman-based DMD branch. 
The input feature map of the BNO network is the spatiotemporal dataset $\boldsymbol{Z} \in \mathbb{R}^{32768 \times 20 \times 1}$, constructed from 20 snapshots of the transverse ($y$-component $v$) velocity field $\boldsymbol{V} \in \mathbb{R}^{256 \times 128 \times 1}$. 
The Koopman operator within BNO is approximated via truncated DMD with $r=12$ modes. 
Broadcasting expands the last dimension of the DMD tensor from $(32768,20,1)$ to $(32768, 20,16)$, since every slice along the third axis (feature dimension) of the DMD output is copied 16 times to match the shape of the Conv2D output. Then, the two tensors are added element-wise.
This is typically used to merge outputs from distinct operator branches---such as convolutional and Koopman-based pathways---thereby combining local and global dynamics into a unified feature space for subsequent processing.
Outputs are merged via an element-wise Add operation, followed by a ReLU and final reconstruction layer.}
\centering
\begin{tabular}{lcccccc}
\hline
\hline
\textbf{Component} & Layer & Kernel & Filters & Input  & Output  & Activation \\
\cmidrule(lr){1-7}
\multicolumn{7}{c}{\textbf{Resolution} $256 \times 128$} \\
\cmidrule(lr){1-7}
 & \textbf{Input} & ... & ... & $(32768,20,1)$ & ... & ...  \\ 
\cmidrule(lr){1-7}
\textbf{CNN Branch}
 & Conv2D & (5,5) & 16 & $(32768,20,1)$ & $(32768,20,16)$ & relu \\
 & Conv2D & (5,5) & 32 & $(32768,20,16)$ & $(32768,20,32)$ & relu \\
 & Conv2D & (5,5) & 16 & $(32768,20,32)$ & $(32768,20,16)$ & relu \\
\cmidrule(lr){1-7}
\textbf{DMD Branch}
 & Reshape & ... & ... & $(32768,20,1)$ & $(32768,20)$ & ... \\
 & DMD & ... & ... & $(32768,20)$ & $(32768,20,1)$ & ... \\
 & Broadcast & ... & ... & $(32768,20,1)$ & $(32768,20,16)$ & ... \\
\cmidrule(lr){1-7}
\textbf{Fusion} 
 & Add & ... & ... & $(32768,20,16)$ & $(32768,20,16)$ & ... \\
 & ReLU & ... & ... & $(32768,20,16)$ & $(32768,20,16)$ & relu \\
 & Conv2D & (5,5) & 1 & $(32768,20,16)$ & $(32768,20,1)$ & linear \\
\hline
\hline
\end{tabular}
\label{tab:bno_parameter}
\end{table}

Table~\ref{tab:bno_parameter} shows the architecture of the BNO model for a single Banach layer at a spatial resolution of $256 \times 128$. 
The network processes spatiotemporal input data by combining convolutional layers with a Koopman-based operator to capture both local and global dynamics.
The Koopman operator provides a linear approximation of the nonlinear spatiotemporal evolution. 
It is implemented using data-driven dynamic mode decomposition (DMD), which captures dominant coherent structures and advances the state forward in time. 
This enables the network to model long-range dependencies and global behavior efficiently. 
The convolutional layers (Conv2D) act as nonlinear correctors.
These layers apply learnable spatial kernels to extract localized features and refine the representation produced by the Koopman operator. 
By projecting the input into a higher-dimensional feature space through an increased number of channels via CNN filters and applying successive transformations, the CNN layers encode fine-scale structures and local interactions.
Activation functions, such as ReLU, are applied between convolutional layers and after the element-wise Add operation to introduce nonlinearity into the operator composition, enhancing the expressiveness of the model. 
Because the outputs of the DMD and Conv2D branches differ in channel dimension, broadcasting is applied to expand the Koopman output across feature channels to match the convolutional representation.
These are then fused via an element-wise Add operation, which integrates global (Koopman) and local (CNN) information into a unified latent space for downstream processing.
The final Conv2D layer with a linear activation reconstructs the output prediction from this combined representation. 

\subsubsection{DMD computation steps}
DMD provides a data-driven technique to approximate the spectral decomposition of a system's evolution. The following outlines the detailed procedure of how to perform DMD analysis. 
Given a system that evolves over time, a sequence of $M-1$ spatiotemporal snapshots of state variables are collected
\begin{align}
	\boldsymbol{Y} &= [\boldsymbol{u}_1, \boldsymbol{u}_2, \cdots, \boldsymbol{u}_{M-1}] \in \mathbb{R}^{N \times (M-1)} \, 
\end{align}
where $\boldsymbol{u}_i \in \mathbb{R}^{N}$ represents the system state at the $i$-th discrete time step. 
Subsequently, the time-shifted snapshots $\boldsymbol{Y}' $ are obtained 
\begin{align}
	\boldsymbol{Y}' &= [\boldsymbol{u}_2, \boldsymbol{u}_3, \cdots, \boldsymbol{u}_{M}] \in \mathbb{R}^{N \times (M-1)} \,.
\end{align}
The goal of DMD is to approximate the best-fit linear operator $\boldsymbol{A} \in \mathbb{C}^{N \times N}$ that governs the evolution 
\begin{align}
	\boldsymbol{Y}' &= \boldsymbol{A} \boldsymbol{Y} \, .
\end{align}
However, it is intractable to analyze the matrix $\boldsymbol{A}$ due to its super-large dimension $M$.
Thus, the reduced matrix $\boldsymbol{\tilde{A}}  \in \mathbb{C}^{r \times r}$ is computed. 

Firstly, the matrix $\boldsymbol{Y}$ is decomposed by employing singular value decomposition (SVD)
\begin{align}
	\boldsymbol{Y} &= \boldsymbol{U} \boldsymbol{\Sigma} \boldsymbol{V}^{*} \, ,
\end{align}
where $\boldsymbol{U} \in \mathbb{C}^{N \times r}$ and $\boldsymbol{V} \in \mathbb{C}^{M \times r}$ contain respectively the left and right singular vectors of $\boldsymbol{Y}$, $\boldsymbol{V}^{*}$ is the complex conjugate transpose of $\boldsymbol{V}$, and $\boldsymbol{\Sigma} \in \mathbb{C}^{r \times r}$ is a diagonal matrix of the singular values. 
Secondly, the reduced matrix $\boldsymbol{\tilde{A}}$ is computed as
\begin{align}
	\boldsymbol{\tilde{A}} &= \boldsymbol{U}^{*} \boldsymbol{Y}' \boldsymbol{V} \boldsymbol{\Sigma}^{-1} \, .
\end{align}
Thirdly, the eigenvalue problem for $\boldsymbol{\tilde{A}}$ is solved to obtain the spectral decomposition
\begin{align}
	\boldsymbol{\tilde{A}} \boldsymbol{W} &= \boldsymbol{W} \boldsymbol{\Lambda} \, ,
\end{align}
where the diagonal matrix $\boldsymbol{\Lambda} = \text{diag}(\lambda)$ contains the discrete time DMD eigenvalues. 
Finally, the DMD modes are given by column vectors of $\boldsymbol{\Phi}$,
\begin{align}
	\boldsymbol{\Phi} &= \boldsymbol{Y}^{'} \boldsymbol{V} \boldsymbol{\Sigma}^{-1} \boldsymbol{W} \, . 
\end{align}
Furthermore, the temporal dynamics can be reconstructed as~\cite{Kutz_2016a}
\begin{align}
	\boldsymbol{y}(x,t) &= \boldsymbol{\Phi} \exp(\boldsymbol{\Omega}t) \boldsymbol{b} \, ,
	 \label{eq:dmd}
\end{align}
where $\boldsymbol{b} = \boldsymbol{\Phi}^{\dag} \boldsymbol{u}_1$  represents the initial conditions. 

\begin{figure}[tbp]
\begin{center}
\includegraphics [width=1\columnwidth]{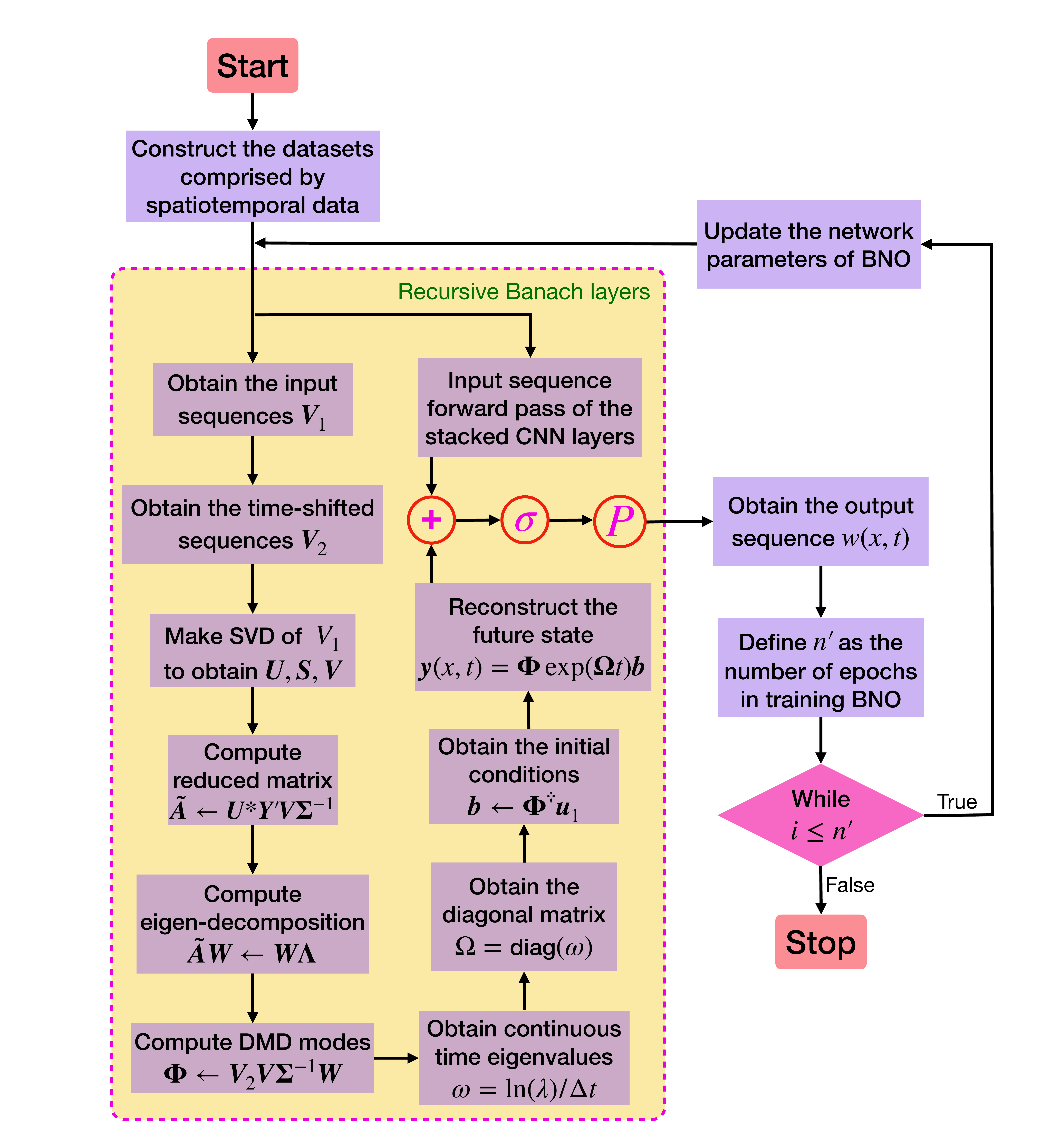}
\end{center}
\caption{Flowchart of the proposed approach for learning the BNO to model the Navier-Stokes equations in an end-to-end, seq2seq manner.}
\label{fig:flowchart}
\end{figure}  

\subsection{Runtime and complexity analysis of BNO components}

In the framework of BNO, the Koopman discrete operator, in concert with the projection operator consisting of stacked CNN operators as defined in Equation~\ref{def:con}, is applied to each data point in the input dataset to forecast the future state in the label dataset. 
An attention-like mechanism is applied through this projection operator to dynamically weigh different parts of the input sequence when making predictions. 
Spatiotemporal features and long-range dependencies are captured by applying convolutional filters over the input sequence through a sliding window mechanism.
The overall procedure for training BNO on the Navier-Stokes equations in an end-to-end, seq2seq manner is illustrated in the flowchart shown in Fig.~\ref{fig:flowchart}.

Since DMD analysis is applied to the input sequence in each Banach layer and the filters in the stacked CNN layers of each Banach layer are shared across different spatiotemporal locations under distinct mesh discretizations, the Banach layers are discretization-invariant. As such, zero-shot super-resolution can be achieved by BNO. 
It is also observed that the computational cost of performing DMD analysis is roughly ten times higher than a forward pass through the stacked CNN layers within each Banach layer.  
The computational complexity of a CNN is determined primarily by the convolutional layers, which dominate the overall workload. 
This complexity depends on factors such as input size, stride, filter size and the number of filters. 
Specifically, a single convolutional layer has complexity $\mathcal{O}(B \cdot O_H \cdot O_W \cdot O_C \cdot K_H \cdot K_W \cdot I_C)$, where $B$ denotes the batch size, $O_H$ and $O_W$ the ouput height and width, $O_C$ the number of output channels, $K_H$ and $K_W$ the kernel height and width, and $I_C$ the number of input channels. 
In contrast, truncated DMD has complexity $\mathcal{O}(Bnmr + Br^3)$, where $n$ and $m$ represent the numbers of spatial features and snapshots, respectively, $r$ is the truncation rank. 
Nonetheless, computational complexity is just one dimension and gives upper bounds, often hiding constant factors, hardware optimization and memory access patterns. 
In practice, DMD is slower than CNN because DMD uses computationally expensive global linear algebra, say SVD and eigendecomposition, over full snapshots sequence. 
In contrast, CNN employs local, parallelizable, GPU-accelerated operations that scale better with large input sizes. 
Moreover, CNNs benefit from mature, optimized libraries and compilers, while DMD in Python is oftentimes CPU-bound and relies on matrix decompositions, which are not GPU-friendly and do not scale easily in parallel. 
Hence, performing DMD analysis and reconstructing temporal dynamics is considerably more computationally expensive than the localized and highly optimized forward pass of stacked CNN operators. 

\section{High-fidelity numerical experiments}
\label{sec:results}
In the experiments, the Navier-Stokes equations were selected as a benchmark problem to evaluate the performance of the proposed neural operator framework. 
The forecasting capability of the BNO for complex mutiscale dynamics was compared against that of the Koopman operator and traditional DNNs. 
Additionally, diverse architectural configurations of the BNO were investigated to assess the impact of design choices on predictive performance. 

\subsection{Simulation setup and computational framework}
To generate the training data, a large eddy simulation (LES) of the unsteady flow over a NACA 4412 airfoil was performed at a chord-based Reynolds number of $Re = \rho_{\infty} U_{\infty}c/\mu = 1.5 \times 10^6$, with a free-stream Mach number of 0.07 and an angle of attack of $\alpha = 14^{\circ}$.
The free-stream conditions are: density $\rho_{\infty}=1.13$ kg/m$^3$, velocity $U_{\infty}=24.45$ m/s, dynamic viscosity $\mu=1.82 \times 10^{-5}$ kg/(m$\cdot$s), and airfoil chord length $c=1.0$ m.

The governing equations are the two-dimensional compressible Navier-Stokes equations, which describe the conservation of mass, momentum and energy: 
\begin{align}
	\partial_t {\rho} + \partial_i({\rho u_{i}}) & = 0 \, ,
 	 \label{eq:continuity1}\\
	 \partial_t({\rho u_{i}}) + \partial_j({\rho u_{i}u_{j}}) & = -\partial_i{p} + \partial_j{\sigma_{ij}} \, ,
	\label{eq:mom} \\
	\partial_t({\rho E}) + \partial_i({\rho E u_i}) & = -\partial_i({p u_i}) + \partial_i({\sigma_{ij} u_i}) - \partial_i{q_i} \, ,
	\label{eq:energy}\,
\end{align}
where $\rho$, $u_i$, and $p$ denote the density, velocity, and pressure, respectively.
The total energy is given by $E = e + \frac{1}{2}\rho u_i u_i$, with internal energy $e = p/(\gamma - 1)$ and specific heat ratio $\gamma = 1.4$.
The viscous stress tensor is defined by $\sigma_{ij} = \mu (\partial_j{u_{i}}+ \partial_i{u_{j}} - \frac{2}{3} \partial_k{u_k} \delta_{ij})$, and the heat flux vector is $q_{i} = -\mu C_p \partial_i{T}/Pr$, where $C_p = 1005$ J/(kg$\cdot$K) is the specific heat at constant pressure, $T$ is the temperature, and $Pr=0.71$ is the molecular Prandtl number.

These equations are discretized using the finite volume method and solved using the rhoPimpleFoam transient solver for compressible flows in OpenFOAM. 
A C-shaped computational mesh is employed, with the airfoil positioned such that its leading edge is located at $x/c=0.15$ and $y/c=0$ in the domain. 

The LES is conducted using the $k$-equation subgrid-scale (SGS) model to represent the effects of usresolved turbulence. 
A first-order implicit Euler scheme is used for temporal discretization to ensure numerical stability during the transient flow evolution.
For spatial discretization, a second-order linear scheme is applied to the diffusive terms to maintain accuracy, while upwind-biased linear schemes were employed for the convective terms associated with velocity, total energy and turbulent kinetic energy.
This combination offers a balance between stability and accuracy in resolving flow features and gradients. 

Freestream velocity boundary conditions (BCs) are applied at both the inlet and outlet of the computational domain. 
A no slip BC is enforced on the airfoil surface. 
For  temperature, inletOutlet BCs are imposed at the inlet and outlet, while a zeroGradient BC is applied on the airfoil wall. 
To simulate varying angles of attack, the flow direction is adjusted by modifying the normal and tangential components of the inlet velocity, rather than rotating the airfoil within the computational domain. 
This approach maintains mesh integrity and simplifies grid generation. 
The fluid properties and flow parameters used in this simulation are selected to align with the experimental conditions reported by Coles and Woodrock~\cite{Coles_1979a}. 
Additional details regarding the simulation configuration can be found in our recent work~\cite{Zhang_2023b, Zhang_2023a}. 

\subsection{Training dataset construction}
The LES field data is interpolated onto a uniform Cartesian grid of size $256 \times 128$, corresponding to a subset of the computational domain defined by $(x/c, y/c) \in [-0.5, 2.5] \times [-0.5, 1.0]$. 
To enhance training efficiency, the flow field data is normalized using the Z-score normalization by subtracting the data mean and dividing by its variance. 
To construct the lower resolution data (\textit{e.g.}, $32 \times 16$), average pooling is applied to the high-resolution ground truth fields ($256 \times 128$).

The transverse velocity field $\boldsymbol{V} \in \mathbb{R}^{256 \times 128 \times 1}$ can be reshaped into a vector $\boldsymbol{u} \in \mathbb{R}^{M'}$, where $M'=32768$.
A data matrix $\boldsymbol{A} \in \mathbb{R}^{M \times N}$ is then constructed by stacking all the velocity components column-wise from the $N' = 4053$ snapshots as
\begin{align}
	\boldsymbol{A} &= [\boldsymbol{u}_1, \boldsymbol{u}_2, \cdots, \boldsymbol{u}_{N'}] \, . 
\end{align}
Each column in this matrix represents a single time snapshot of the reshaped transverse velocity field.

A subset of these snapshots is then used to construct the training dataset.
Specificallly, each data point consists of $n \in \mathbb{N}$ consecutive snapshots sampled with a temporal stride of $k \Delta t$ with $k \in \mathbb{N}$.
There is a total of $m \in \mathbb{N}$ data points in the training dataset. 
Thus, the input of the dataset is constructed as follows
\begin{equation*}
\begin{aligned}
	u_{input} = \{\{ u^{k\Delta t}, u^{2k\Delta t}, u^{3k\Delta t}, \cdots, u^{(n-1)k\Delta t}, u^{nk\Delta t} \} \, , \\
	 \{ u^{2k\Delta t}, u^{3k\Delta t}, u^{4k\Delta t}, \cdots, u^{nk\Delta t}, u^{(n+1)k\Delta t} \} \, , \\
	 \cdots \cdots \cdots \cdots \cdots \cdots \cdots \cdots \cdots \cdots \\
	  \{ u^{mk\Delta t}, u^{(m+1)k\Delta t}, u^{(m+2)k\Delta t}, \cdots, u^{(m+n-2)k\Delta t}, u^{(m+n-1)k\Delta t} \}\} \, , 
\end{aligned}
\end{equation*}
while the label of the dataset is obtained by shifting each state variable $sk$ time steps with $s \in \mathbb{N}$ in the input dataset 
\begin{equation*}
\begin{aligned}
	u_{label} = \{\{ u^{(s+1)k\Delta t}, u^{(s+2)k\Delta t}, u^{(s+3)k\Delta t}, \cdots, u^{(s+n-1)k\Delta t}, u^{(s+n)k\Delta t} \} \, , \\
	 \{ u^{(s+2)k\Delta t}, u^{(s+3)k\Delta t}, u^{(s+4)k\Delta t}, \cdots, u^{(s+n)k\Delta t}, u^{(s+n+1)k\Delta t} \} \, , \\
	 \cdots \cdots \cdots \cdots \cdots \cdots \cdots \cdots \cdots \cdots \\
	  \{ u^{(s+m)k\Delta t}, u^{(s+m+1)k\Delta t}, u^{(s+m+2)k\Delta t}, \cdots, u^{(s+m+n-2)k\Delta t}, u^{(s+m+n-1)k\Delta t} \}\} \, , 
\end{aligned}
\end{equation*}
where $u \in \mathbb{R}^{N}$ represents the system state at discrete time steps. 
Here, the tuple $(n,k,m,s)^T$ can be varied as part of a parametric study in future work. 
Furthermore, 70\% and 30\% of the 500 data points in the dataset are used for training and validation, respectively. 

The mean squared error (MSE) is employed as the loss function for convergence of the BNO network
\begin{align}
	\text{Loss} & = \frac{1}{N_B \cdot N_x \cdot N_t} \sum^{N_B}_{n=1} \sum^{N_x}_{i=1} \sum^{N_t}_{j=1} \Bigl(\boldsymbol{X}_{n,i,j} - \boldsymbol{\tilde{X}}_{n,i,j} \Bigl) \, ,
 	 \label{eq:mse}\,
\end{align}
where $N_B=10$ is the mini-batch size, $N_x=32768$ is the number of spatial points, $N_t=20$ is the number of stacked time frames, $\boldsymbol{X}_{n,i,j}$ is the ground truth, and $\boldsymbol{\tilde{X}}_{n,i,j}$ is the corresponding prediction from the BNO network. 
$N_x$ and $N_t$ are the height and width of the input data.

\subsection{Benchmarking: one-layer BNO versus DMD and CNN}
To evaluate the effectiveness of the proposed BNO architecture, a controlled benchmarking study is conducted using a single Banach layer as defined in Equations~\ref{def:con} and \ref{def:kdo}.
This initial setup serves to isolate and evaluate the core benefits of incorporating a Koopman-based operator learning strategy compared to baseline methods.

The performance of the BNO is compared with that of DMD and a CNN model composed of stacked convolutional layers for predicting the transverse velocity field. 
The CNN model corresponds to the embedded CNN component within the BNO framework, but it is trained independently without incorporating the Koopman discrete operator, as illustrated in the BNO architecture in Fig.~\ref{fig:architecture}.

\subsubsection{Training setup}
The models are trained using the Adam optimizer with a batch size of 10 and a piecewise constant learning rate schedule: an initial learning rate of $10^3$ for the first 1500 iterations, reduced to $10^{-4}$ for the following 1000 iterations, and further decreased to $10^{-5}$ thereafter. 
A single Banach layer is used, with a truncation rank of 12 for the DMD.
Each Conv2D layer is configured with a kernel size of $5 \times 5$.
ReLU activation is applied between CNN layers, except for the output layer, where a linear activation is used.
Training is conducted over $10^4$ to $3 \times 10^4$ epochs on a single Apple M4 Max GPU (40-core with 128GB unified memory) using the Tensorflow framework.

The Koopman operator within the BNO framework is approximated via truncated DMD with $r=12$ modes, while a consistent architecture is adopted for the CNN components across all experiments.
The dataset consists of spatiotemporal solutions at varying grid resolutions, with lower resolution inputs generated by applying average pooling to high-resolution ground truth data.
All models are trained on the same dataset, specified by $(n,k,m,s)^T = (20,2,350,1)^T$.

\begin{figure}[tbp]
\begin{center}
\includegraphics [width=.7\columnwidth]{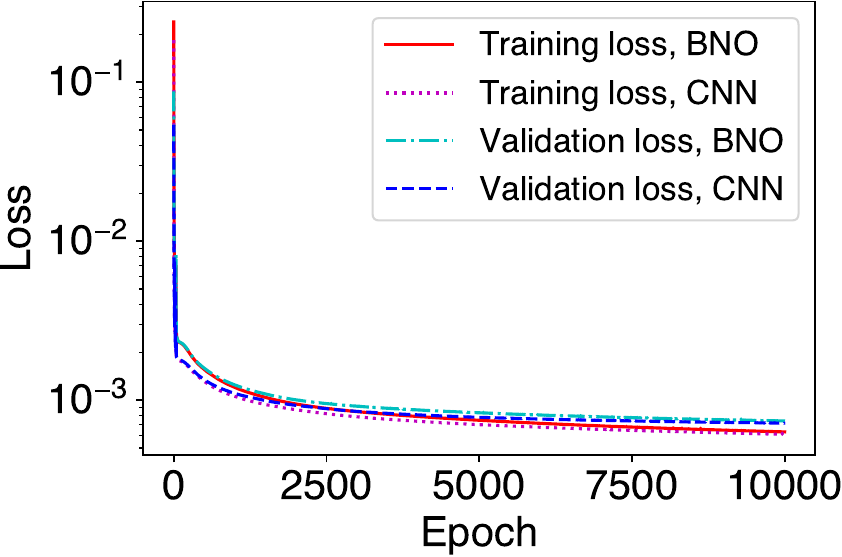}
\end{center}
\caption{Training and validation loss histories of BNO and CNN models for the transverse velocity field evaluated at a resolution of $256 \times 128$.}
\label{fig:loss_128by256}
\end{figure} 

\begin{figure}[tbp]
\begin{center}
\includegraphics [width=1\columnwidth]{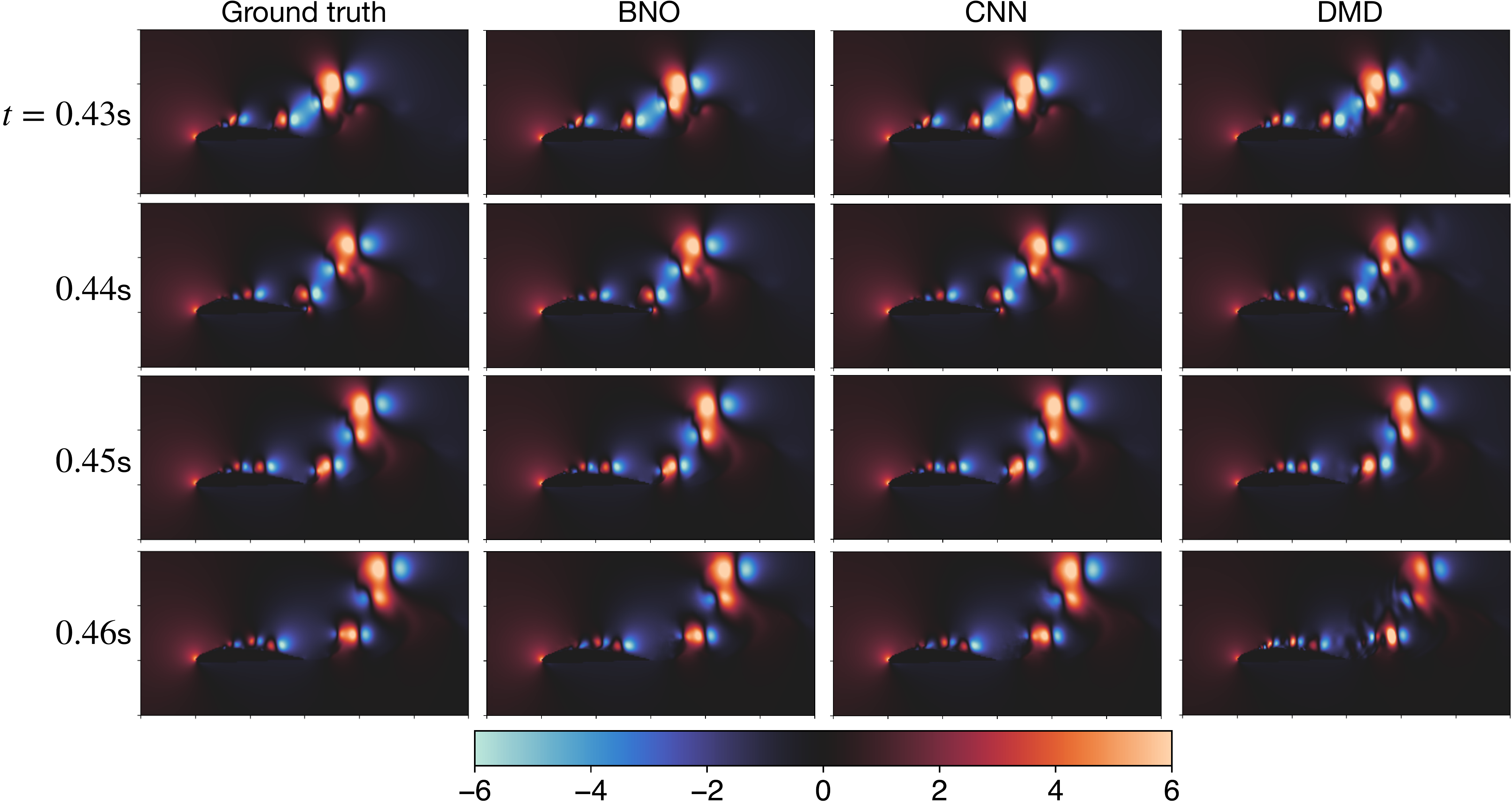}
\end{center}
\caption{Spatiotemporal evolution of the transverse velocity field evaluated at a resolution of $256 \times 128$. The first and fourth columns show the ground truth and DMD prediction, respectively; the second and third columns present the learned interpolations by the BNO and CNN models. In this setting, the BNO is constructed with a single Banach layer. Furthermore, one-step-ahead prediction on the training dataset is applied to capture the spatiotemporal dynamics for both the BNO and CNN models.}
\label{fig:compare1}
\end{figure}  

\begin{figure}[tbp]
\begin{center}
\includegraphics [width=1\columnwidth]{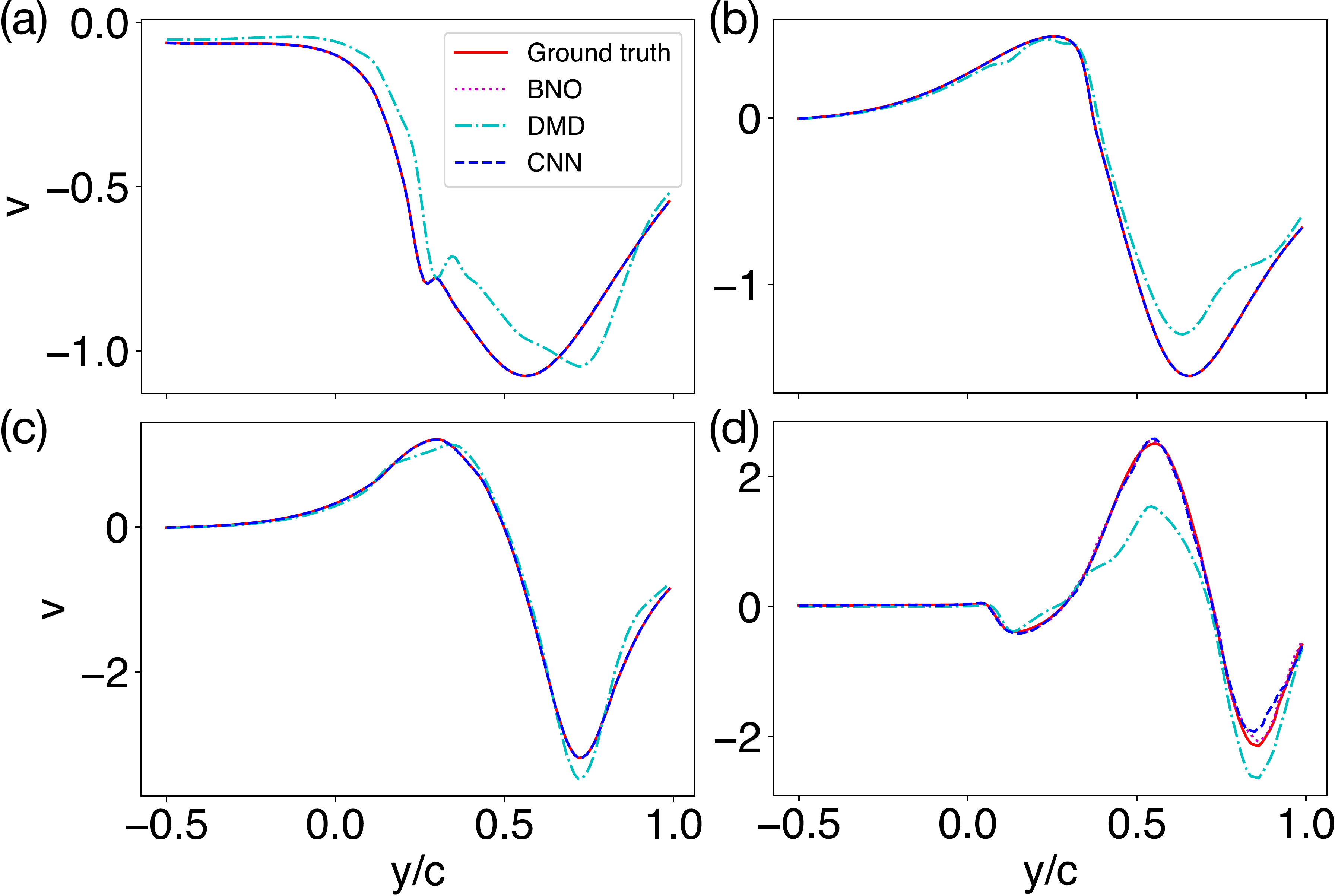}
\end{center}
\caption{Comparison of the temporal evolutions of the transverse velocity profiles predicted by the BNO, DMD and CNN models against the ground truth at $x/c=1.8$ and distinct time instants (a) $t=0.43$s, (b) $0.44$s, (c) $0.45$s, and (d) $0.46$s. A one-step-ahead prediction strategy is applied to capture the spatiotemporal dynamics.
The results are obtained using the transverse velocity field evaluated at a resolution of $256 \times 128$.}
\label{fig:line1}
\end{figure}  

\begin{figure}[p]
\begin{center}
\includegraphics [width=1\columnwidth]{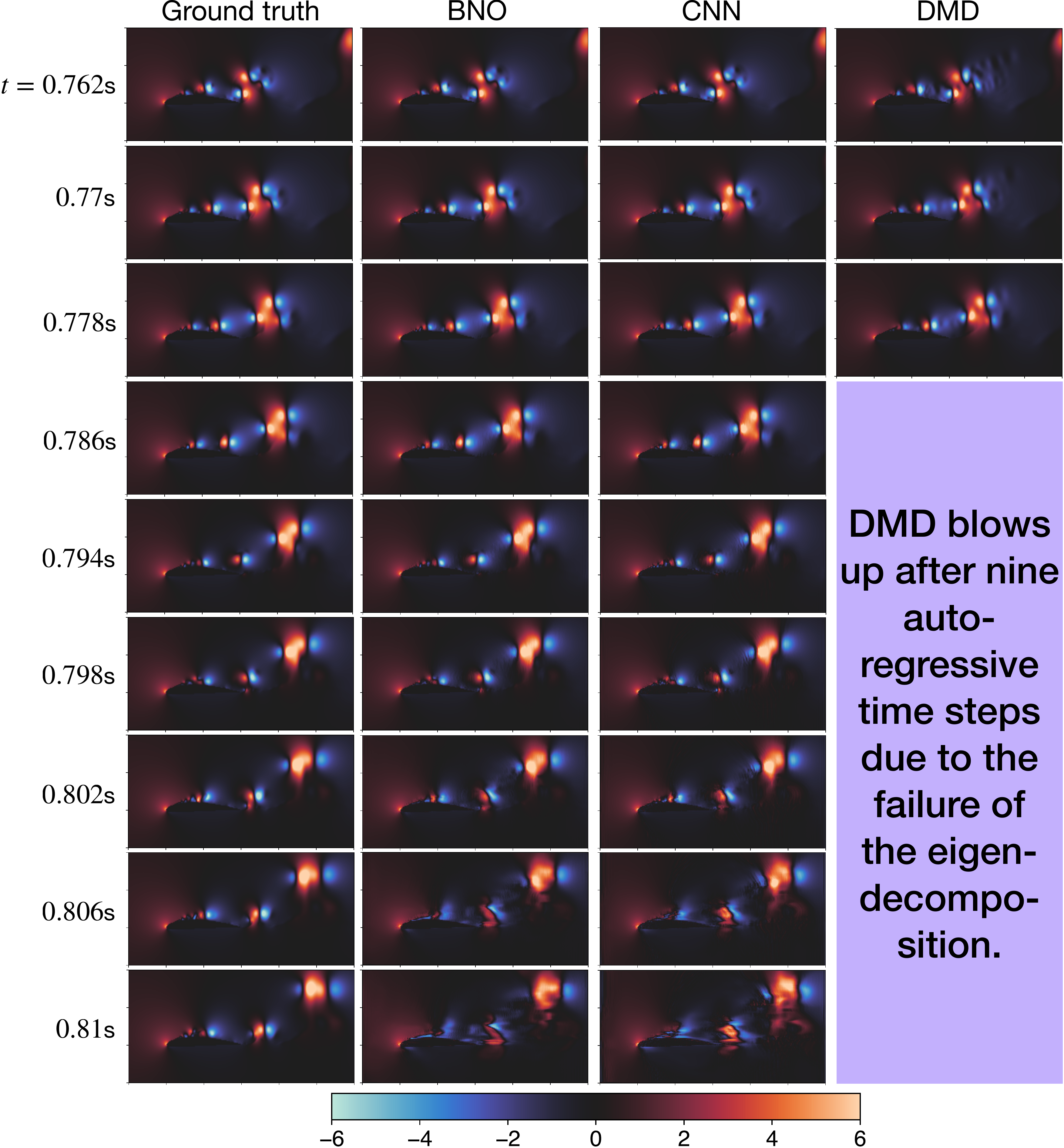}
\end{center}
\caption{Spatiotemporal evolution of the transverse velocity field evaluated at a resolution of $256 \times 128$ using autoregressive prediction on the training dataset. In this scenario, the BNO is constructed using a single Banach layer.}
\label{fig:compare2}
\end{figure}  

\begin{figure}[tbp]
\begin{center}
\includegraphics [width=1\columnwidth]{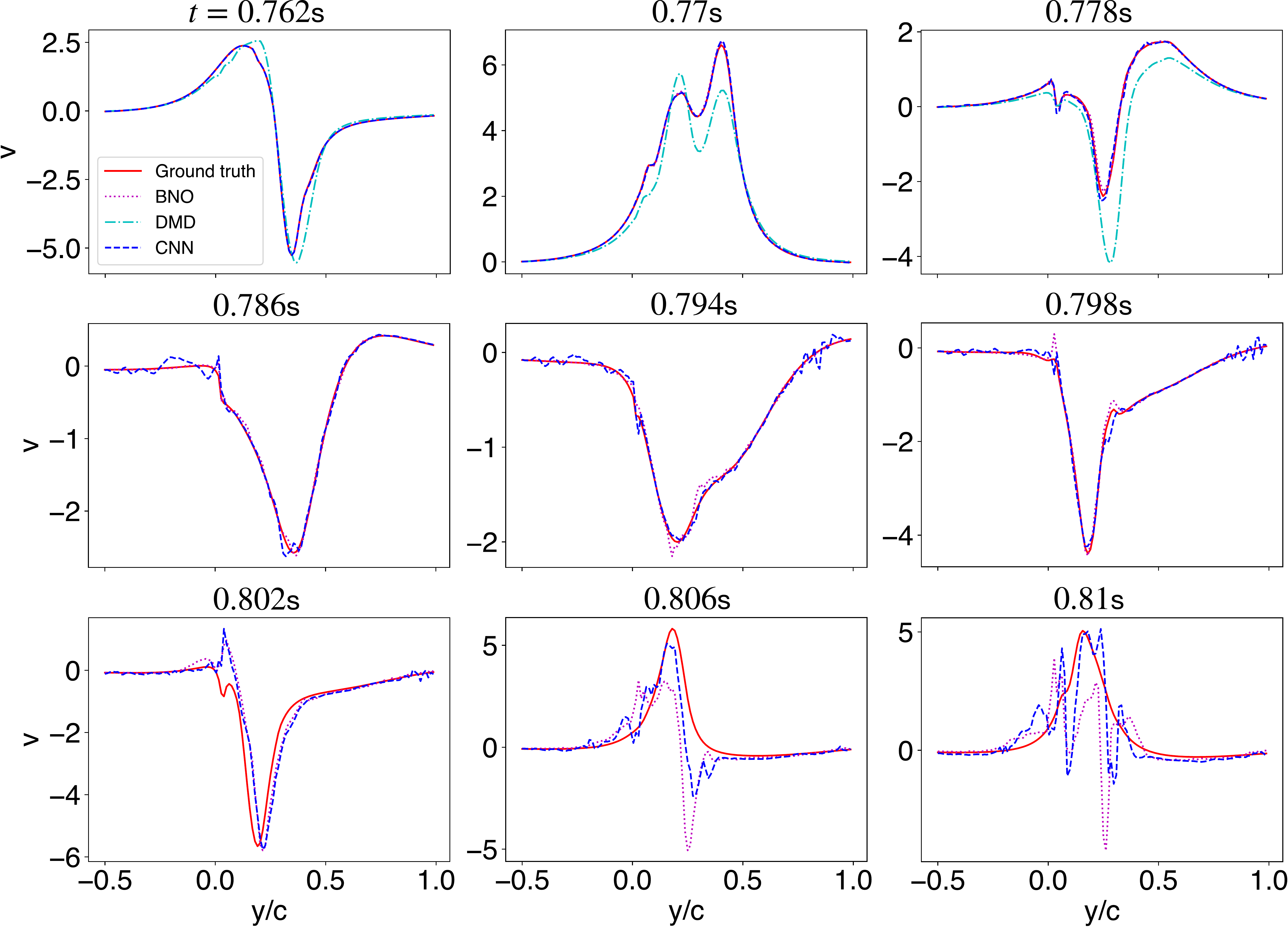}
\end{center}
\caption{Comparison of the temporal evolutions of the transverse velocity profiles predicted by the BNO, DMD and CNN models against the ground truth at $x/c=1.2$ and selected time instants. For $t \ge 0.786$s, DMD becomes unstable, resulting in divergence of the predicted solution.
The results are obtained using the transverse velocity field evaluated at a resolution of $256 \times 128$.}
\label{fig:line2}
\end{figure}  

\begin{figure}[p]
\begin{center}
\includegraphics [width=1\columnwidth]{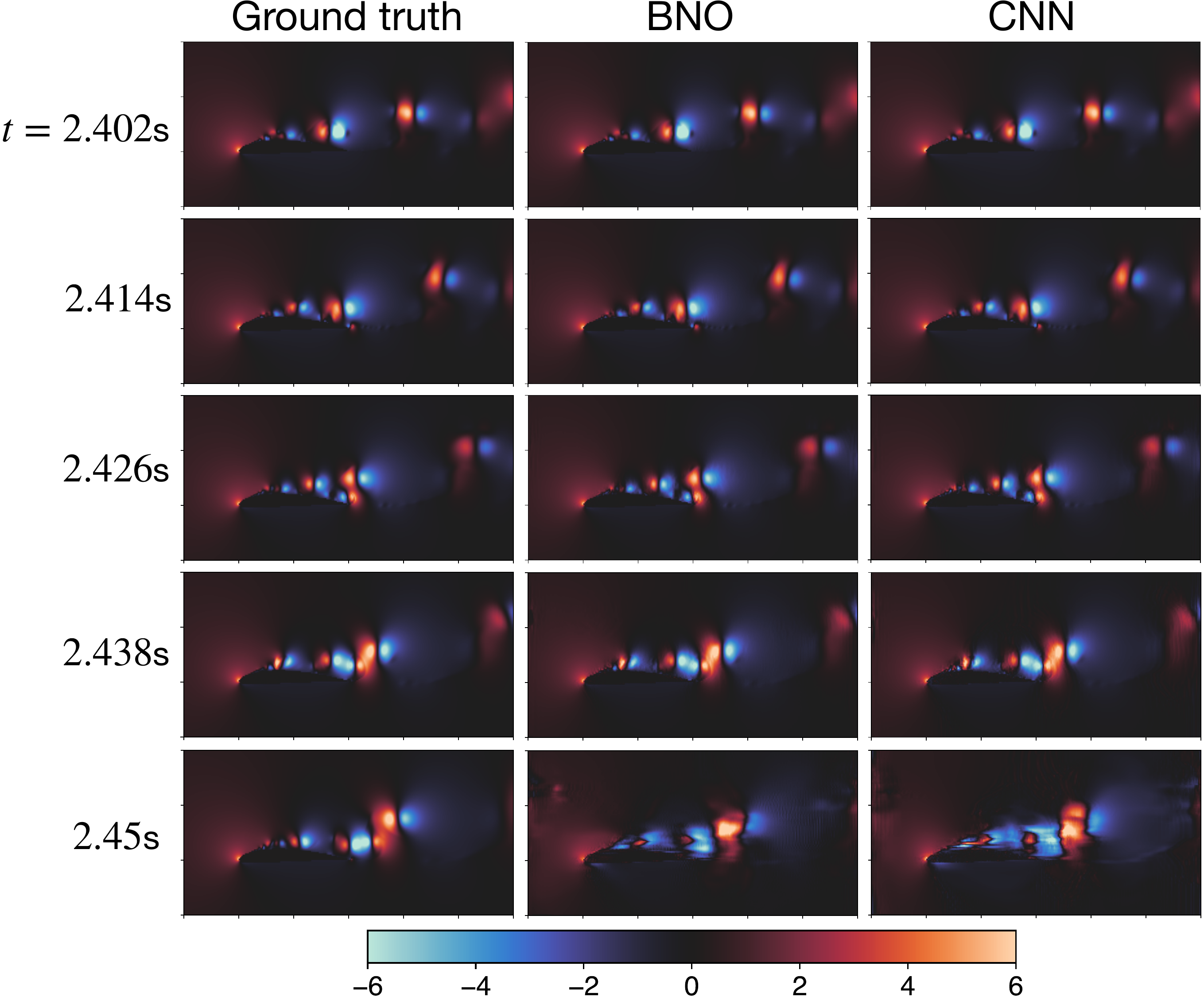}
\end{center}
\caption{Spatiotemporal evolution of the transverse velocity field evaluated at a resolution of $256 \times 128$ using autoregressive prediction on the unseen test dataset.}
\label{fig:compare3}
\end{figure}  

\begin{figure}[tbp]
\begin{center}
\includegraphics [width=1\columnwidth]{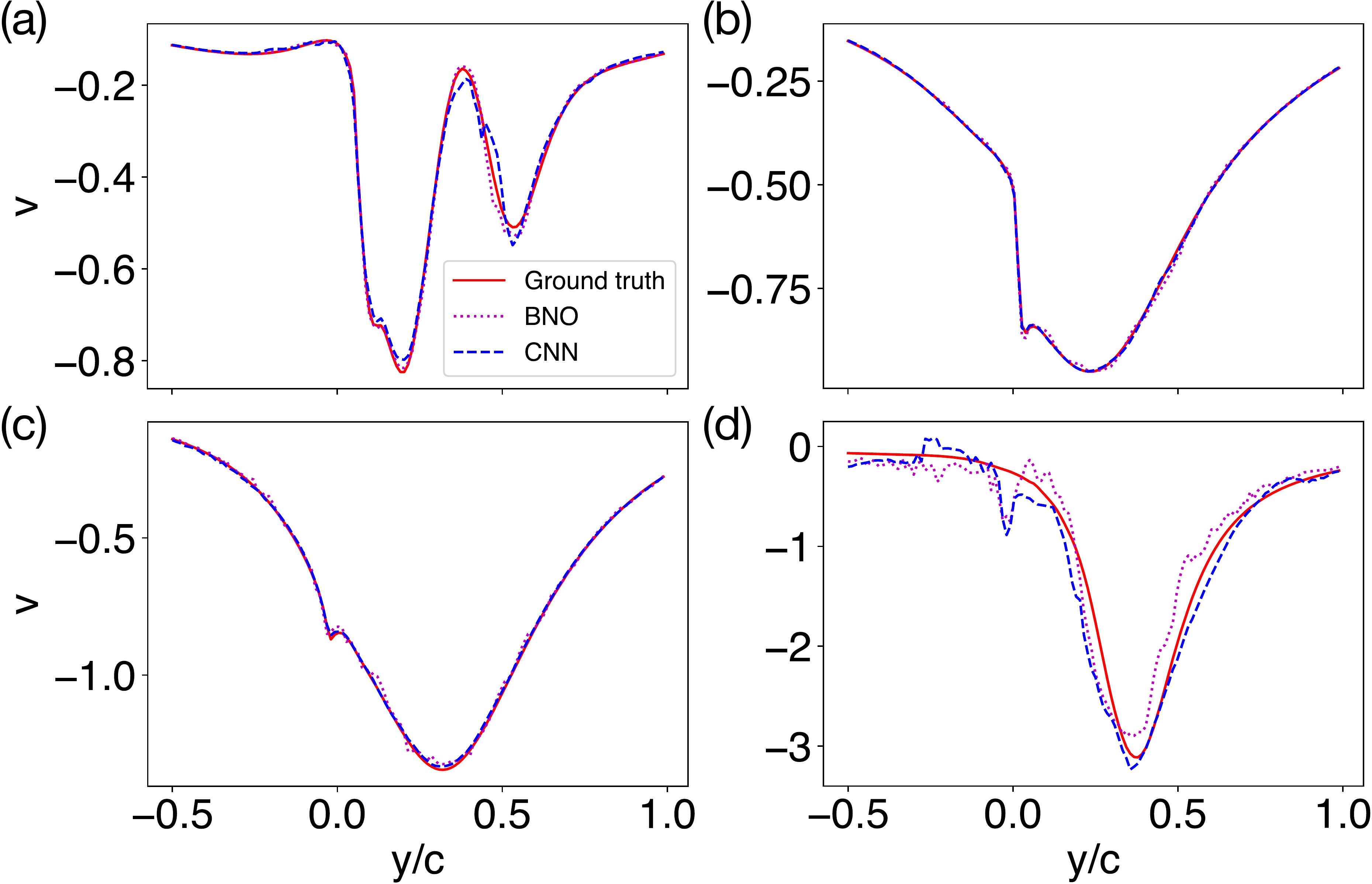}
\end{center}
\caption{Comparison of the temporal evolutions of the transverse velocity profiles predicted by the BNO and CNN models against the ground truth at $x/c=1.6$ and distinct time instants (a) $t=2.414$s, (b) $2.426$s, (c) $2.438$s, and (d) $2.45$s using autoregressive prediction on the unseen test dataset.
The results are obtained using the transverse velocity field evaluated at a resolution of $256 \times 128$.}
\label{fig:line3}
\end{figure}  

\begin{table}[h!]
\setlength{\tabcolsep}{.33em}
\centering
\caption{Comparison of the final epoch loss (MSE) between BNO with two distinct configurations and CNN across different resolution grids after $10^4$ training iterations.}
\label{tab:loss_comparison}
\begin{tabular}{lcccc}
\hline  
\hline
Model & Resolution & Training Loss & Validation Loss & Overfitting Gap \\
\hline  
1-layer BNO & $256 \times 128$ & $6.31 \times 10^{-4}$ & $7.391 \times 10^{-4}$ & $1.081 \times 10^{-4}$ \\
CNN             & $256 \times 128$ & $6.087 \times 10^{-4}$ & $7.161 \times 10^{-4}$ & $1.074 \times 10^{-4}$ \\
1-layer BNO & $64 \times 32$     & $3.151 \times 10^{-4}$ & $4.113 \times 10^{-4}$ & $9.62 \times 10^{-5}$ \\
CNN             & $64 \times 32$     & $2.176 \times 10^{-4}$ & $3.772 \times 10^{-4}$ & $1.596 \times 10^{-4}$ \\
1-layer BNO & $32 \times 16$     & $2.297 \times 10^{-4}$ & $3.71 \times 10^{-4}$ & $1.413 \times 10^{-4}$ \\
CNN             & $32 \times 16$     & $1.271 \times 10^{-4}$ & $3.483 \times 10^{-4}$ & $2.212 \times 10^{-4}$ \\
2-layer BNO & $32 \times 16$     & $4.422 \times 10^{-4}$ & $6.434 \times 10^{-4}$ & $2.01 \times 10^{-4}$ \\
\hline  
\hline
\end{tabular}
\end{table}

\subsubsection{Convergence behavior and overfitting analysis}
Fig.~\ref{fig:loss_128by256} presents training and validation loss histories for the transverse velocity field at a resolution of $256 \times 128$.
The overall trends in the training and validation loss histories of BNO and CNN are similar. 
Although both training and validation losses converge to comparable values, the CNN achieves slightly lower loss values than the BNO, indicating better convergence during training.
The overfitting gap, measured by the difference between the training and validation losses at the last epoch (quantified via the mean squared error, MSE), is approximately equal for both models. 
See Table~\ref{tab:loss_comparison} for a detailed comparison. 

The performance difference can be attributed to architectural differences.
CNN has a well-established architecture with standardized convolutional layers and training routines. 
These networks are relatively shallow and do not involve complex operator composition, rendering them easier to optimize with gradient-based methods. 
BNO, on the other hand, integrates both Koopman operator and CNN within a unified framework. 
This hybrid structure introduces additional complexity in terms of operator learning in infinite-dimensional function spaces and nonlinear coupling between the Koopman component and the CNN corrector. 
CNN typically has fewer constraints and more direct parameterization, allowing the model to efficiently learn local spatial features.
In contrast, BNO involves functional mappings and operator-based learning, which can be more difficult to train due to its higher sensitivity to initialization, more intricate loss landscapes, and possible mismatches in representation capacity between the Koopman and CNN components. 
Furthermore, training stability and gradient flow are generally more reliable in CNNs due to their well-established architectures and fewer dependencies across distant layers or iterations.
In BNO, improper coordination between the Koopman and CNN components or errors in the Koopman estimation can lead to unstable gradients or slow convergence. 
Additionally, CNN often fits the training data more easily due to their localized receptive fields and inductive biases, leading to better convergence metrics, \textit{i.e.}, lower training loss. 
BNO, designed for better generalization and extrapolation in complex spatiotemporal dynamics, intentionally trades off convergence speed and final training loss to ensure robustness and long-term predictive stability.  
While CNN converges better during training due to their simpler and more direct structure, this does not necessarily imply superior performance in extrapolation or generalization. 
BNO may converge more slowly but is often more accurate in capturing global dynamics over time, particularly in nonlinear systems such as fluid flows. 

\subsubsection{One-step forecasting accuracy}
The BNO and CNN models predict each subsequent time step in an autoregressive fashion, using its previous prediction as input. 
As shown in Fig.~\ref{fig:compare1}, BNO demonstrate superior performance in inferring the velocity field at various time instants, as evidenced by visual inspection. 
In addition, both the BNO and CNN models outperform DMD in the one-step-ahead prediction task. 
Nevertheless, DMD is unable to capture the complex nonlinear dynamical features across all time steps. 
In particular, it fails to predict the coherent structures in the shear layer, wake region and near-field of the airfoil at $t=0.46$s. 
Furthermore, BNO and CNN achieve comparable accuracy in capturing the spatiotemporal evolution of the coherent structures around the airfoil. 

Fig.~\ref{fig:line1} presents a quantitative assessment of the predictive abilities of BNO, DMD and CNN by comparing the temporal evolution of velocity line profiles against the ground truth. 
As shown in the figure, the velocities predicted by BNO and CNN align well with the ground truth across all time steps. 
Notably, at $t=0.46$s, BNO demonstrates superior accuracy compared to CNN, which shows a noticeable misalignment in the second valley of the velocity proflie.
In contrast, DMD exhibits significant deviations from the ground truth at the earlier and later time instants of $t=0.43$s and $0.46$s, though its predictions show improved agreement at $t=0.45$s. 
BNO utilizes the Koopman operator to model spatiotemporal dynamics, while the integrated CNN module acts as a corrector that enhances prediction accuracy. 
This hybrid architecture enables BNO to outperform both CNN and DMD in capturing complex flow behaviors. 

\subsubsection{Autoregressive forecasting and stability}
To evaluate long-term predictive stability, each model is tested under nine-step autoregressive forecasting on the training dataset.
Fig.~\ref{fig:compare2} shows that for $t \le 0.778$s, DMD fails to accurately reconstruct the velocity field in the shear layer and the wake region of the airfoil. 
After $t = 0.786$s, DMD becomes numerically unstable due to ill-conditioned eigen-decomposition, resulting in divergence of the velocity field.
In contrast, both BNO and CNN demonstrate consistently higher accuracy than DMD across the first nine autoregressive time steps. 
Moreover, for $t \ge 0.786$s, BNO outperforms CNN in capturing the complex nonlinear spatiotemporal dynamics in the wake region of the airfoil. 
At $t=0.802$s, both BNO and CNN exhibit reduced accuracy in resolving the velocity field close to the trailing-edge of the airfoil. 
This trend continues for $t \ge 0.806$s, where both models experience a general decline in prediction accuracy across the entire flow field. 
Notably, by $t \ge 0.806$s, neither model is able to accurately reconstruct the velocity distribution in the vicinity of the leading-edge. 

Fig.~\ref{fig:line2} presents corresponding velocity line profiles.
For $t \le 0.778$s, DMD demonstrates significantly lower accuracy compared to both BNO and CNN. 
While BNO and CNN exhibit comparable performance for $t \le 0.77$s, BNO surpasses CNN and aligns more closely with the ground truth in the range $0.778 \le t \le 0.798$s.   
For $t \ge 0.802$s, both models deviate, but CNN's predictions exhibit higher variance and oscillations.
This can be attributed to BNO's Koopman-based global evolution, which provides regularization by embedding linear dynamics aligned with the system's global modes.

\subsubsection{Extrapolation to unseen test dataset}
To evaluate generalization beyond the training domain, all models are tested on an unseen test dataset.
Fig.~\ref{fig:compare3} illustrates that, for $t \le 2.438$s, both the BNO and CNN exhibit similar predictive performance, closely aligning with the ground truth. 
At $t = 2.45$s, noticeable spatial errors emerge in both models.

In Fig.~\ref{fig:line3}, BNO predictions align better with the ground truth at $t = 2.414$s.
At $t = 2.426$s and $2.438$s, both BNO and CNN achieve good agreement with the ground truth. 
Nevertheless, by $t = 2.45$s, both models exhibit noticeable spatial misalignment. 

\subsubsection{Summary}
The benchmarking study confirms that the BNO architecture---while more challenging to train---achieves superior generalization and long-term stability compared with CNN and DMD.
The embedded Koopman operator enhances its ability to capture global flow dynamics, making BNO a robust model for nonlinear spatiotemporal systems under both interpolation and extrapolation conditions.

\subsection{Grid independence study}

\begin{figure}[tbp]
\begin{center}
\includegraphics [width=.7\columnwidth]{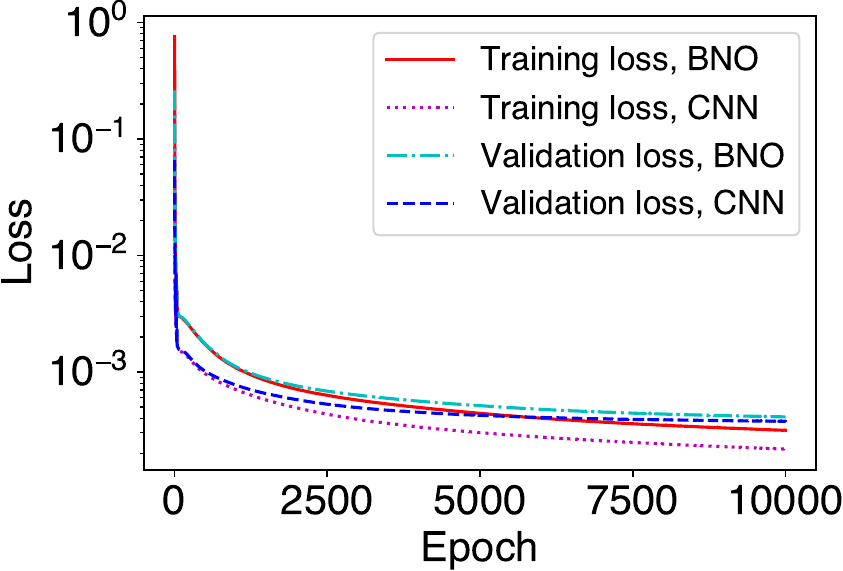}
\end{center}
\caption{Training and validation loss histories of BNO and CNN models for the transverse velocity field evaluated at a resolution of $64 \times 32$.}
\label{fig:loss_32by64}
\end{figure} 

\begin{figure}[p]
\begin{center}
\includegraphics [width=1\columnwidth]{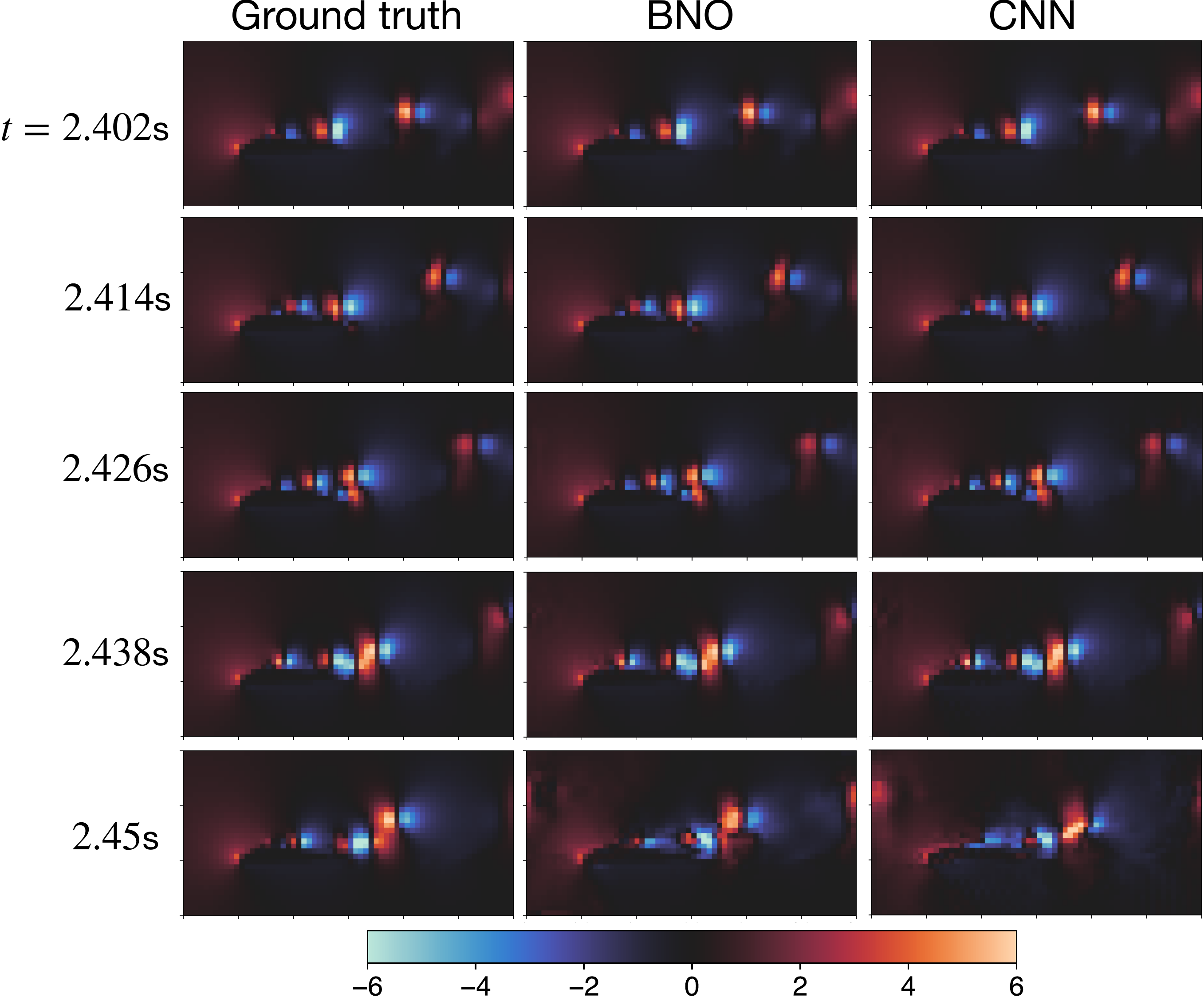}
\end{center}
\caption{Spatiotemporal evolution of the transverse velocity field evaluated at a resolution of $64 \times 32$ using autoregressive prediction on the unseen test dataset.}
\label{fig:compare_grid}
\end{figure}  

\begin{figure}[tbp]
\begin{center}
\includegraphics [width=1\columnwidth]{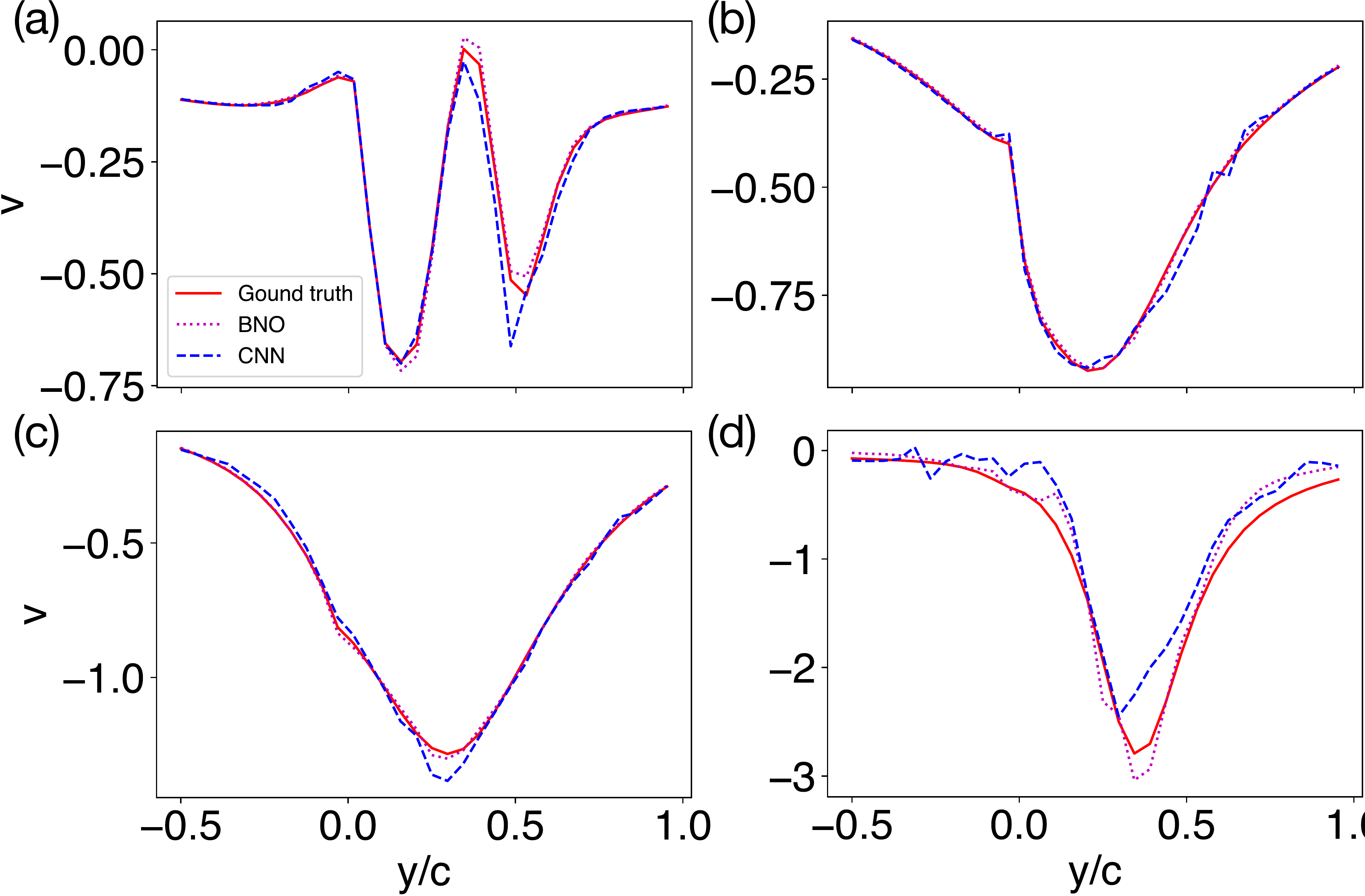}
\end{center}
\caption{Comparison of the temporal evolutions of the transverse velocity profiles predicted by the BNO and CNN models against the ground truth at $x/c=1.6$ and distinct time instants (a) $t=2.414$s, (b) $2.426$s, (c) $2.438$s, and (d) $2.45$s using autoregressive prediction on the unseen test dataset.
The results are obtained using the transverse velocity field evaluated at a resolution of $64 \times 32$.}
\label{fig:line_grid}
\end{figure}  

A formal grid independence study was conducted to evaluate the sensitivity of the model to changes in spatial resolution. 
The BNO model was independently trained on datasets with different resolutions, including $32 \times 16$, $64 \times 32$, and $256 \times 128$. 
Performance was assessed across these resolutions using consistent evaluation metrics. 
Fig.~\ref{fig:loss_32by64} shows the training and validation loss histories for the transverse velocity field at a resolution of $64 \times 32$.
Compared to BNO, the CNN exhibits a more pronounced overfitting trend, as evidenced by a larger overfitting gap in Table~\ref{tab:loss_comparison}. 
While the CNN attains lower loss values during training---indicating faster convergence---the BNO demonstrates stronger generalization capability.

To assess generalization beyond the training domain, all models are evaluated on an unseen test dataset.
As shown in Fig.~\ref{fig:compare_grid}, for $t \le 2.438$s, both BNO and CNN perform comparably, closely matching the ground truth. 
However, for $t \ge 2.45$s, noticeable spatial errors arise in both models.
The BNO model maintains superior performance, accurately capturing the velocity field, particularly in the shear layer and wake regions where nonlinear dynamics are dominant.

Fig.~\ref{fig:line_grid} presents a quantitative comparison of velocity profiles at $x/c=1.6$ across four time instants.
BNO predictions consistently align more closely with the ground truth across all evaluated times.
At $t = 2.426$s and $2.438$s, the BNO shows excellent agreement with the ground truth. 
By $t = 2.45$s, although both models exhibit spatial discrepancies, BNO continues to outperform CNN in accuracy.
Notably, the BNO model trained on $64 \times 32$ achieves even better accuracy than the one trained on the high-resolution $256 \times 128$ data, highlighting the stability and resolution robustness of the learned operator. 
These results support the grid independence of the proposed approach in a data-driven learning context.

\subsection{Generalization via zero-shot super-resolution}
In this experiment, the zero-shot super-resolution capability of the BNO configured with a single Banach layer is evaluated. 
Specifically, its ability to infer high-resolution flow fields is assessed using the same learned network parameters trained exclusively on coarse-resolution input data, without any additional fine-tuning or retraining. 

\begin{figure}[tbp]
\begin{center}
\includegraphics [width=.7\columnwidth]{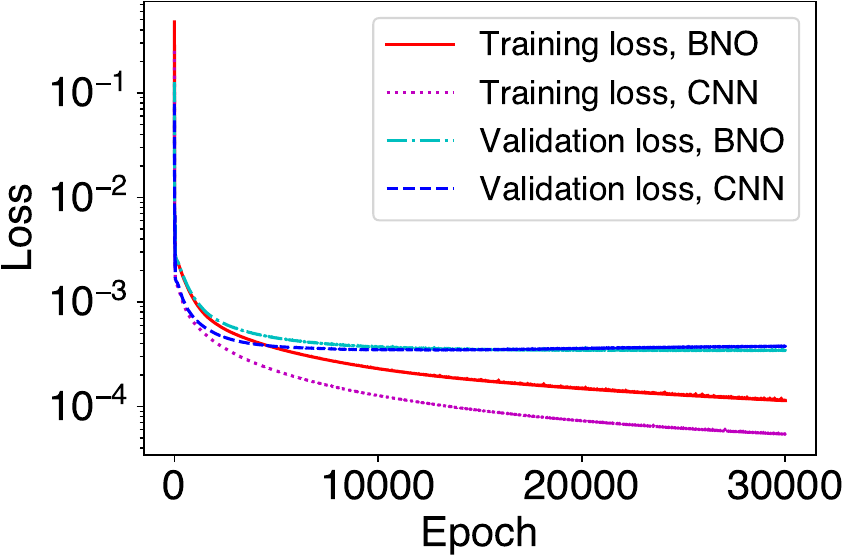}
\end{center}
\caption{Training and validation loss histories of BNO and CNN models for the transverse velocity field evaluated at a resolution of $32 \times 16$.}
\label{fig:loss_16by32}
\end{figure} 

\begin{figure}[p]
\begin{center}
\includegraphics [width=1\columnwidth]{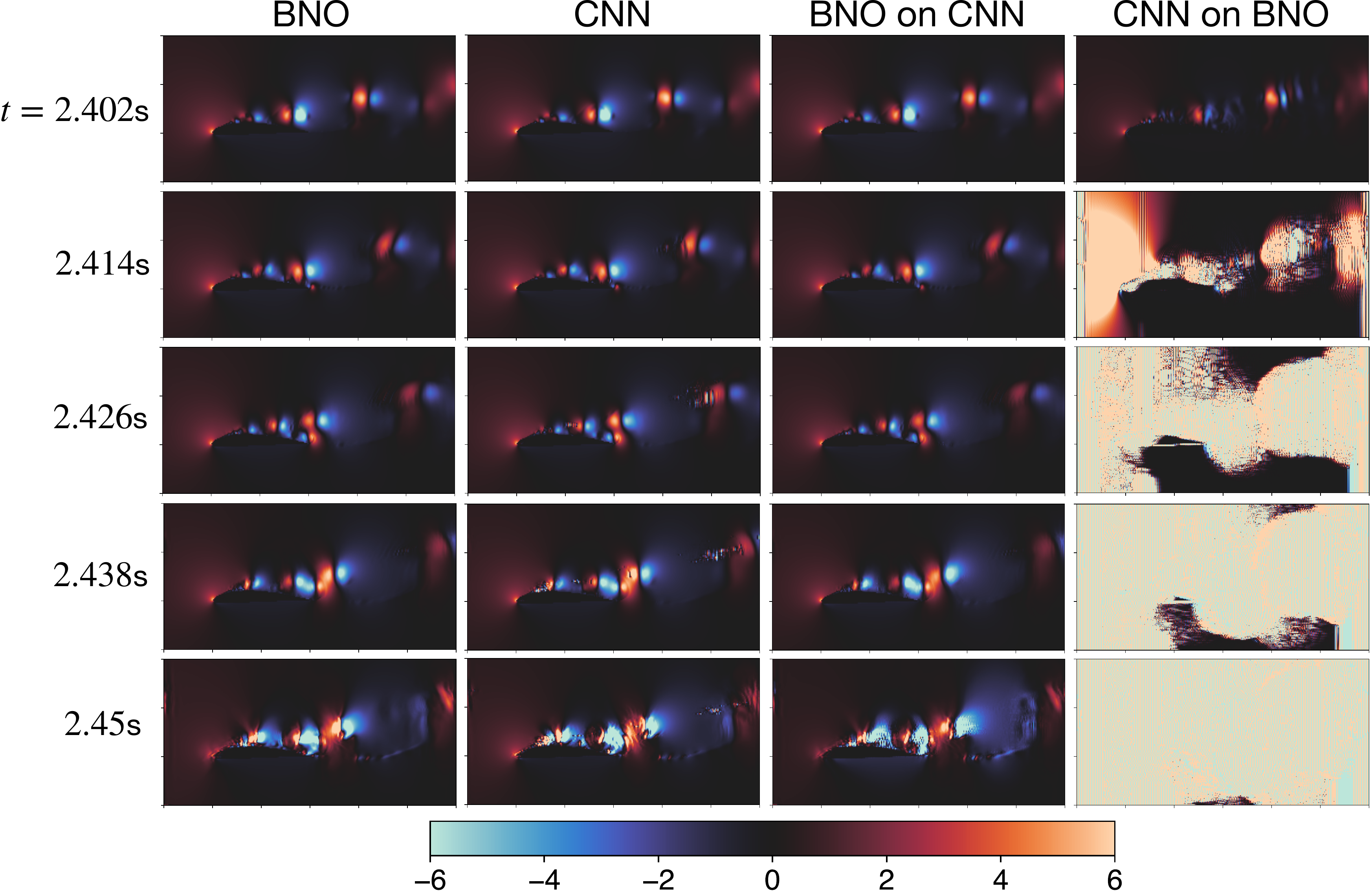}
\end{center}
\caption{Zero-shot super-resolution of the transverse velocity field evaluated at a resolution of $256 \times 128$, using network parameters trained exclusively on coarse-resolution data ($32 \times 16$). The third and fourth columns depict the reconstructed velocity fields generated by applying the learned BNO and CNN network parameters to the CNN and BNO architectures, respectively.}
\label{fig:compare4}
\end{figure} 

\begin{figure}[tbp]
\begin{center}
\includegraphics [width=.95\columnwidth]{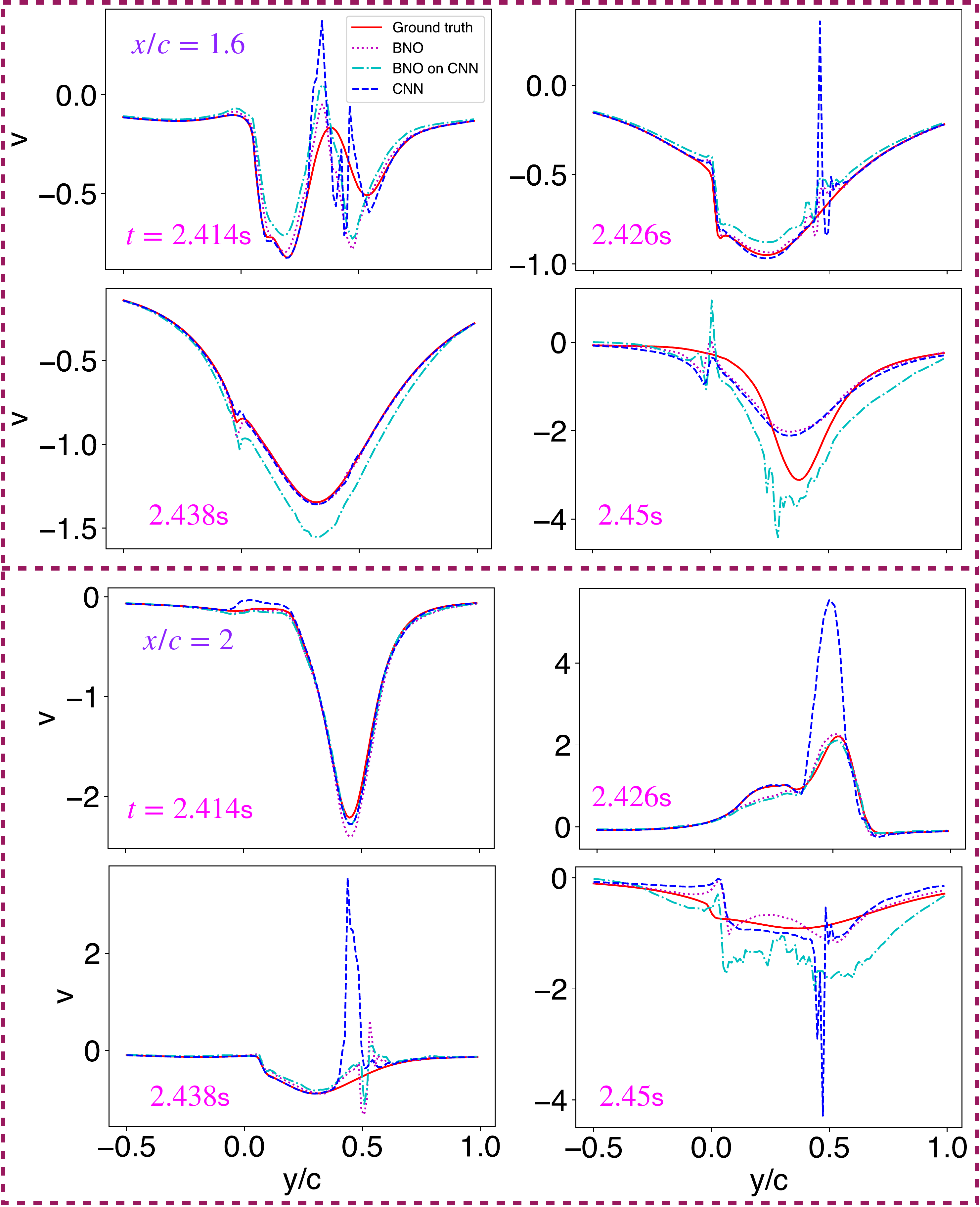}
\end{center}
\caption{Comparison of the temporal evolutions of the transverse velocity profiles predicted by the BNO and CNN models against the ground truth at $x/c=1.6$ (top panel) and $x/c=2$ (bottom panel). The models are trained on coarse-resolution data ($32 \times 16$) and evaluated on high-resolution data $256 \times 128$ without retraining}.
\label{fig:line4}
\end{figure}  

\subsubsection{Training performance on coarse data}
Fig.~\ref{fig:loss_16by32} shows the training and validation loss histories at a coarse resolution of $32 \times 16$.
The CNN displays a more pronounced overfitting trend compared to BNO, which is quantitatively supported by the overfitting gap in Table~\ref{tab:loss_comparison}.
This is consistent with CNN's behavior that CNN learns local patterns using convolutional filters, which renders them effective at fitting detailed features in the training data. 
However, this also means CNN is more likely to memorize specific spatial features from the training set, especially when autoregressively applied over time. 
In contrast, BNO, by incorporating a Koopman operator, is designed to model global spatiotemporal relationships more robustly. 
This acts as a regularizing effect, preventing the model from simply memorizing local fluctuations. 

CNN has a strong inductive bias toward local correction, which works well on training data but may fail to generalize to unseen long-term dynamics. 
BNO's architecture, especially its operator-theoretic component, emphasizes global consistency and structure, leading to better generalization even with higher training error. 
CNN often optimize to a very low training loss, which can be misleading---they may fit the noise or small-scale structures that do not generalize. 
BNO, while possibly converging to a slightly higher training loss, tends to focus on coherent dynamical structures, improving extrapolation. 
In autoregressive prediction, \textit{i.e.} step-by-step time evolution, CNN can accumulate error faster due to overfitting to training trajectories. 
BNO handles long-term dependencies more gracefully, reducing drift and maintaining physical consistency over time. 

\subsubsection{Super-resolution across resolutions}
Zero-shot super-resolution tests a model's ability to infer high-resolution outputs from coarse-trained parameters, without exposure to the target resolution during training. This is particularly valuable in scenarios where high-resolution data is scarce or expensive to generate and applications that require model portability across different grid scales. 
In this context, BNO is trained exclusively on a $32 \times 16$ low-resolution dataset, yet it is able to infer $256 \times 128$ high-resolution transverse velocity fields without seeing any high-resolution data, thereby demonstrating zero-shot super-resolution capability. 
While the FNO has been shown to perform zero-shot super-resolution by training on $64 \times 64$ resolution data and transferring to $256 \times 256$ spatial resolution~\cite{Li_2020a}, the proposed BNO model achieves successful super-resolution transfer from an even coarser resolution, highlighting its enhanced generalization ability across a larger resolution gap. 
Fig.~\ref{fig:compare4} presents a qualitative comparison between the zero-shot super-resolution capabilities of BNO and CNN. 
At $t = 2.402$s, both models achieve comparable predictive accuracy, closely aligning with the ground truth. 
However, for $t \ge 2.414$s, the BNO demonstrates superior zero-shot super-resolution performance, capturing the velocity field with higher accuracy than CNN, particularly in the wake regions. 

At $t = 2.45$s, the predictive performance of both models begins to degrade, though BNO still maintains relatively better alignment with the ground truth. 
Moreover, the transferability of network parameters was evaluated by applying the learned BNO network parameters within the CNN architecture.
As shown in Fig.~\ref{fig:compare4}, zero-shot super-resolution can be achieved even when the BNO-trained parameters are transferred to the CNN network, indicating the robustness and generalization capability of BNO-trained representations. 
The resulting predictive performance is comparable to that of the original BNO model. 
In contrast, applying the learned CNN network parameters within the BNO architecture fails to produce accurate high-resolution reconstructions, revealing weaker generalization capacity and limited transferability of CNN-trained weights to the BNO framework.

\subsubsection{Quantitative evaluation}
Fig.~\ref{fig:line4} provides a quantitative comparison of the temporal evolution of velocity profiles at $x/c = 1.6$ and $x/c = 2$.
At $x/c = 1.6$, BNO shows closer alignment with the ground truth for $t \le 2.426$s, outperforming both CNN and the cross-architecture variant.
Furthermore, even under cross-architecture application---where BNO-trained parameters are deployed within the CNN framework---zero-shot super-resolution remains successful, with the resulting predictions closely aligning with the ground truth and surpassing those of the CNN in accuracy. 
At $t = 2.438$s, both BNO and CNN yield predictions that closely match the ground truth and outperform those produced by the cross-architecture application, which exhibits noticeable deviations and reduced accuracy in reconstructing the velocity field. 
Nevertheless, by $t = 2.45$s, all models---including BNO, CNN, and the cross-architecture application---show noticeable spatial misalignment with the ground truth. 

Further downstream at $x/c = 2$ and $t = 2.414$s, all models demonstrate high predictive accuracy, with their predictions closely aligning with the ground truth. 
For $t \ge 2.426$s, the BNO and cross-architecture application continue to produce predictions that remain in relatively good agreement with the ground truth. 
In contrast, the CNN exhibits a marked decline in predictive accuracy, with increasingly pronounced discrepancies in reconstructing the velocity field, particularly in regions of complex flow dynamics. 

\subsubsection{Summary}
These results demostrate that BNO achieves strong generalization through its hybrid operator-CNN structure, enabling effective zero-shot super-resolution across an eightfold resolution gap.
Its ability to generalize spatially---without retraining---suggests strong potential for practical applications where training data are scarce, high-resolution measurements are costly, or grid transferability is required.
The architectural flexibility and parameter robustness further underscore the advantage of embedding operator-theoretic priors within deep learning models.

\subsection{Deeper architectures: two Banach layers}

\begin{figure}[tbp]
\begin{center}
\includegraphics [width=.7\columnwidth]{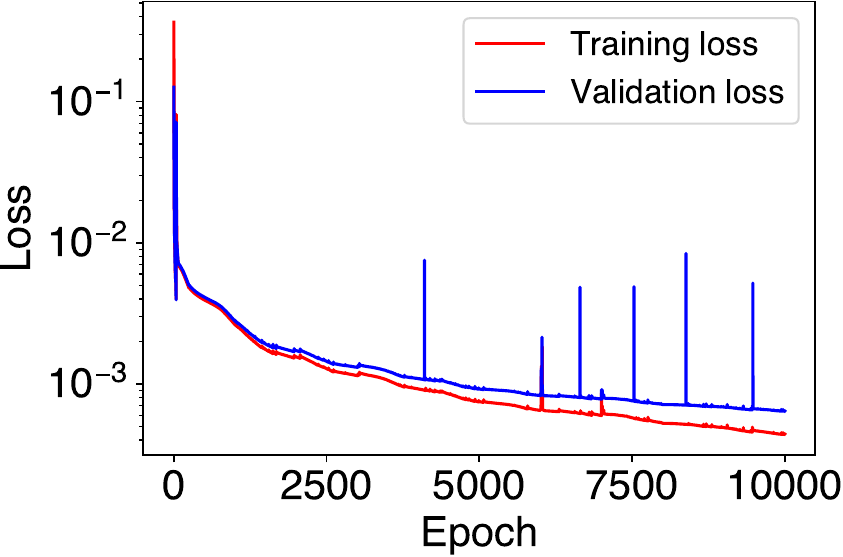}
\end{center}
\caption{Training and validation loss histories of BNO model for the transverse velocity field evaluated at a resolution of $32 \times 16$ employing two Banach layers.}
\label{fig:loss_16by32_2layers}
\end{figure} 

\begin{figure}[p]
\begin{center}
\includegraphics [width=1.\columnwidth]{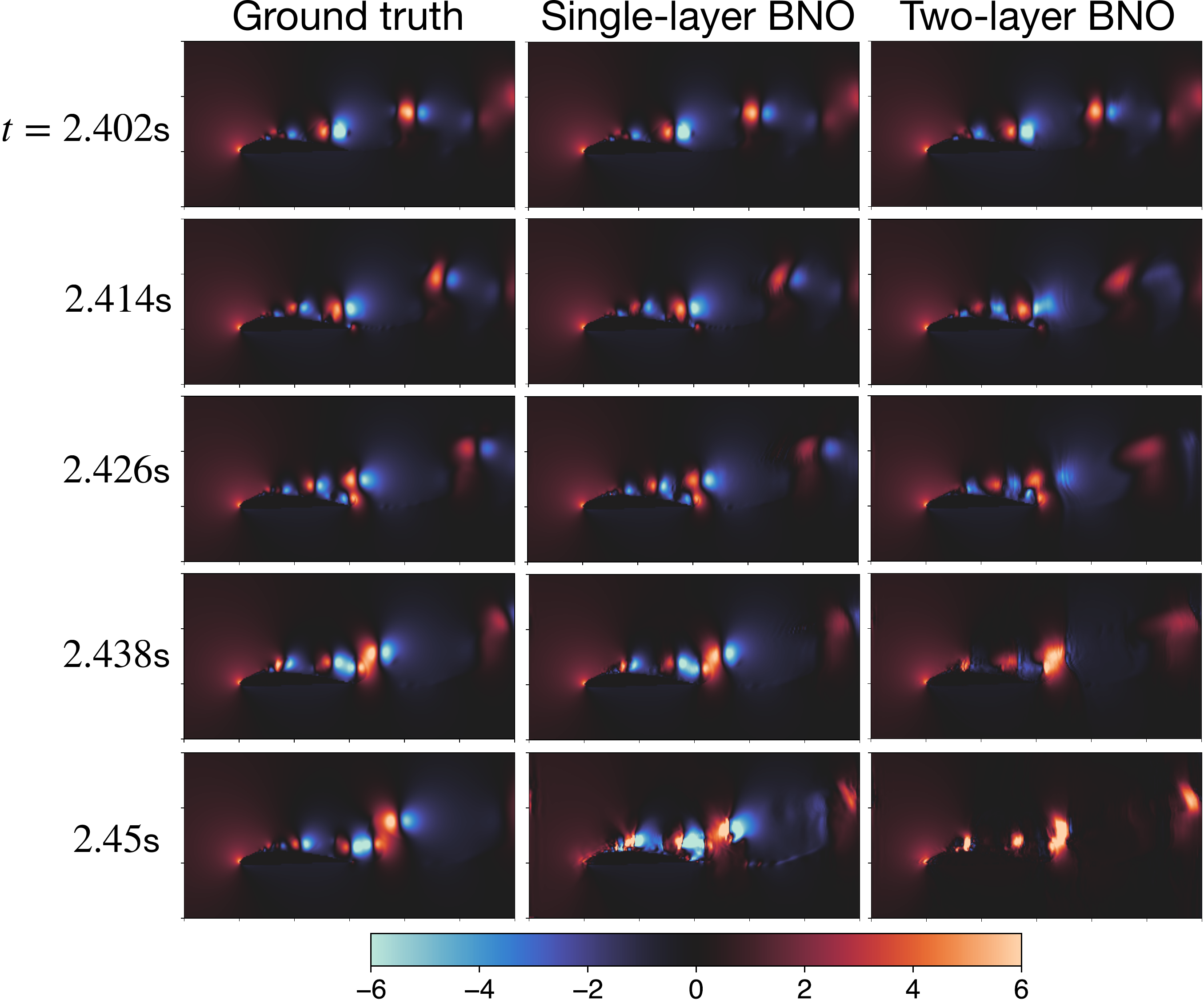}
\end{center}
\caption{Zero-shot super-resolution of the transverse velocity field, evaluated at a high-resolution of $256 \times 128$, using network parameters trained exclusively on coarse-resolution data ($32 \times 16$). Two architectural configurations of the BNO model are compared: one with a single Banach layer and the other with two Banach layers.}
\label{fig:compare5}
\end{figure} 

\begin{figure}[tbp]
\begin{center}
\includegraphics [width=1\columnwidth]{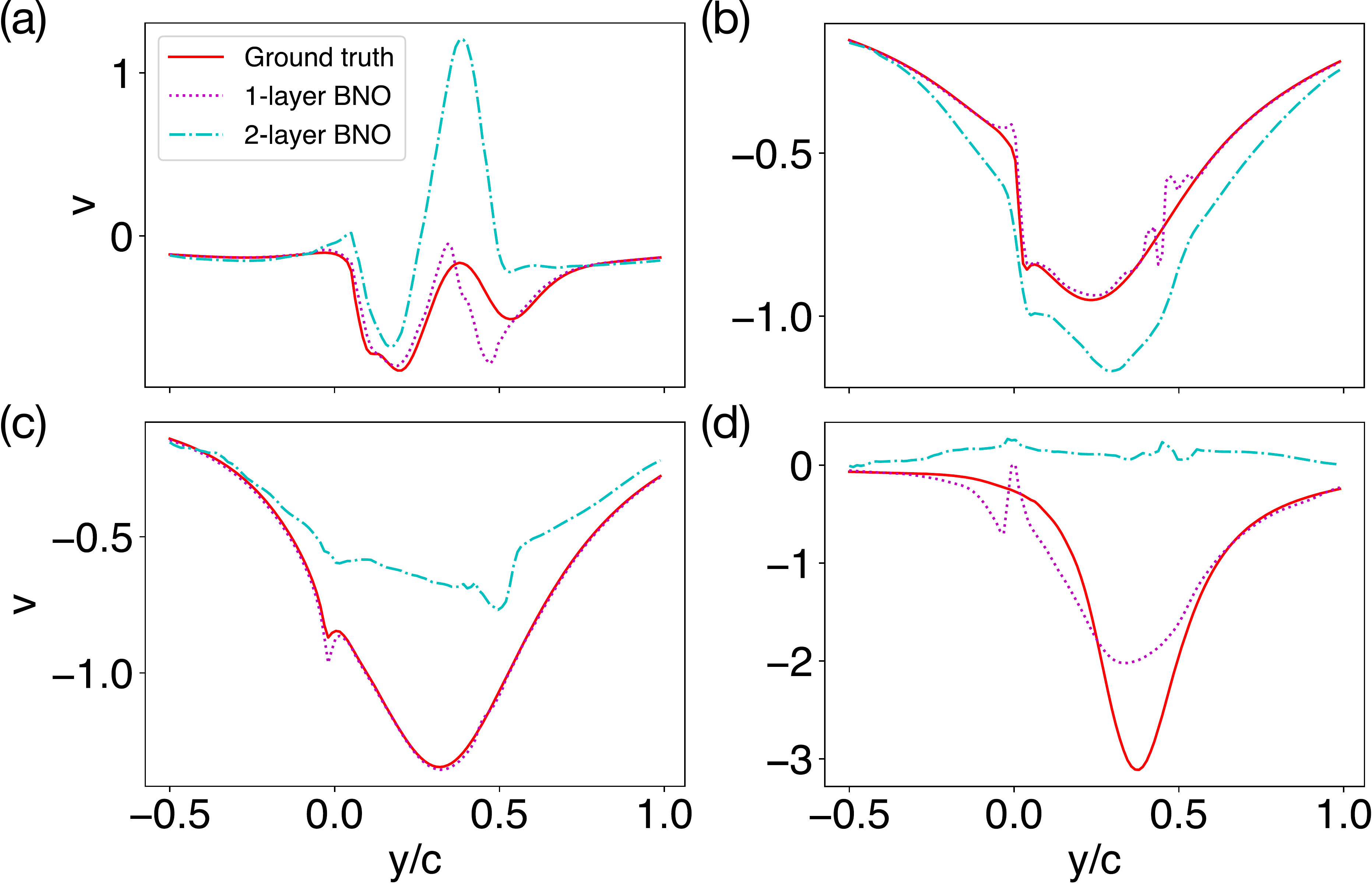}
\end{center}
\caption{Comparison of the temporal evolution of transverse velocity profiles at $x/c=1.6$ across four distinct time instants: (a) $t=2.414$s, (b) $2.426$s, (c) $2.438$s, and (d) $2.45$s. Predictions are generated by the BNO model and compared against the ground truth. The models are trained on coarse-resolution data ($32 \times 16$) and evaluated on high-resolution data ($256 \times 128$) without retraining, under two architectural configurations: one with a single Banach layer and the other with two Banach layers.}
\label{fig:line5}
\end{figure} 

In this subsection, the predictive performance of the BNO architecture is evaluated when extended to a deeper configuration with two Banach layers.
The aim is to assess whether increased model depth improves the ability to capture complex spatiotemporal dynamics or instead introduces instability and performance degradation.

\subsubsection{Training instability and numerical challenges}
As shown in Fig.~\ref{fig:loss_16by32_2layers}, the training and validation losses for the two-layer BNO configuration, evaluated at $32 \times 16$ resolution, exhibit greater variance and higher final error compared to the single-layer model (see Table~\ref{tab:loss_comparison}).
The training process is constrained to $10^4$ epochs, beyond which numerical issues---including breakdown of the  Cholesky decomposition---become increasingly likely.

The observed numerical instability arises from the failure of Cholesky decomposition in the eigendecompostion process.
When employing multiple Banach layers, the computation becomes increasingly unstable, and the Cholesky decomposition step within the eigendecompostion process is prone to failure, potentially causing the training of the BNO to break down. 
Additionally, the validation loss history exhibits more pronounced fluctuations compared to the training loss, indicating increased sensitivity and instability in the model's generalization performance. 
As shown in Table~\ref{tab:loss_comparison}, the overfitting gap at the last epoch is larger than that of the single-layer BNO evaluated at a resolution of $32 \times 16$.  
Cholesky decomposition requires the input matrix to be symmetric positive definite (SPD), and its failure typically indicates the matrix has lost this property. 
As depth increases, maintaining this SPD property becomes more difficult due to the following factors:
\begin{enumerate}
  \item Accumulated numerical error across layers: Each Banach layer involves operator learning steps that form Gram matrices like $X^{T}X$. As layers deepen, small numerical errors can accumulate, degrading the conditioning of these matrices. Eventually, the matrix may acquire near-zero or negative eigenvalues, violating the SPD condition. 
  \item Rank deficiency in intermediate representations: If the representations produced by intermediate layers become low-rank, the resulting matrix $A = X^{T}X$ becomes singular or near-singular ($\det(A) \approx 0$), making it non-SPD and incompatible with Cholesky decomposition. 
  \item Ill-conditioned matrices: Deep BNOs can suffer from poor conditioning of intermediate matrices, where even small round-off errors in floating-point arithmetic cause matrices that should be SPD to appear non-SPD, leading to numerical failures. 
  \item Gradient instability during training: Deep BNOs are more susceptible to exploding or vanishing gradients during backpropagation. These unstable gradients can lead to parameter updates that corrupt intermediate representations, producing matrices that are numerically invalid for Cholesky decomposition. 
\end{enumerate}
In summary, Cholesky decomposition fails in deep BNOs due to the emergence of non-SPD matrices caused by accumulated numerical instability, rank deficiency, ill-conditioning, and gradient-related issues during training. 
These challenges highlight the need for careful regularization, improved numerical conditioning, or alternative decomposition strategies when extending BNO architectures to greater depth, which will be relegated to the future work. 

\subsubsection{Super-resolution performance comparison}
Fig.~\ref{fig:compare5} presents a qualitative comparison of the zero-shot super-resolution capabilities of the BNO architecture configured with a single Banach layer versus two Banach layers. 
At early times ($t = 2.402$s), both configurations yield visually similar reconstructions that closely match the ground truth.
However, as time progresses ($t \ge 2.414$s), the single Banach layer configuration demonstrates superior zero-shot super-resolution performance, capturing the velocity field with higher accuracy than the two-layer configuration, particularly in both the shear layer and wake regions, where nonlinear dynamics dominate. 
By $t \ge 2.438$s, the two-layer model fails to capture key flow structures, while the single-layer model maintains reasonable accuracy.

Fig.~\ref{fig:line5} presents a quantitative comparison using velocity profiles at $x/c=1.6$ across four time instants.
For $t \le 2.438$s, the single-layer configuration demonstrates superior accuracy in capturing the velocity field, showing closer alignment with the ground truth and outperforming the two-layer configuration. 
By $t = 2.45$s, both configurations display noticeable spatial misalignment relative to the ground truth, but the single-layer configuration retains better predictive fidelity. 

\subsubsection{Summary}
These results indicate that deeper BNO configurations, despite having increased expressive capacity, suffer from reduced generalization and numerical instability.
The single-layer BNO strikes a more effective balance between complexity and robustness, achieving better predictive performance and stable training under zero-shot super-resolution settings.
Future work will explore architectural modifications and regularization strategies to enable deeper operator-learning models without compromising numerical stability and generalization.

\section{Conclusions}
\label{sec:conclusion}
This work introduces the Banach Neural Operator (BNO), a novel neural operator framework that integrates Koopman operator theory with deep neural networks to model complex, nonlinear dynamical systems through mappings between infinite-dimensional Banach spaces.
To address the limitations of traditional neural operators---particularly their inability to capture inter-input dependencies due to the absence of attention mechanisms---an end-to-end sequence-to-sequence formulation is developed to learn dominant spatiotemporal patterns with both interpretability and predictive power.
An attention-like mechanism is implemented via a projection operator composed of stacked CNN layers, enabling the model to dynamically weigh different parts of the input sequence when making predictions without relying on explicit query-key-value computations. 

BNO further leverages Koopman analysis to provide a structured, physically motivated decomposition of system dynamics---yielding interpretable insights into the dominant modes, temporal evolution, and correction mechanisms. 
This interpretability is particularly valuable for modeling complex nonlinear systems, where traditional black-box neural networks may achieve high accuracy but remain opaque. 
Moreover, the iterative nature of BNO allows for a natural formulation as a recurrent neural network, enabling parameter sharing across layers and combining the expressiveness of neural operators with the parameter efficiency of recurrent models.  

Comparative evaluations demonstrate that while CNN exhibits faster convergence and achieves lower training loss due to its simpler architecture and well-established optimization procedures, the BNO framework ultimately delivers superior performance in forecasting complex spatiotemporal dynamics, particularly in long-term autoregressive prediction tasks. 
Despite a more challenging training process, the integration of a Koopman operator within the Banch space architecture enables BNO to maintain predictive stability and robustness over extended time horizons.
Quantitative and qualitative analyses across both training and unseen datasets consistently show that BNO outperforms DMD and offers competitive or improved accuracy relative to CNNs. 
These results validate the effectiveness of the hybrid operator-learning approach in BNO and highlights its potential for applications requiring reliable extrapolation in fluid dynamics modeling. 

The study also highlights the remarkable zero-shot super-resolution capability of BNO, particularly when configured with a single Banach layer. 
Despite being trained solely on coarse-resolution data, BNO effectively infers high-resolution flow fields without any additional fine-tuning, demonstrating strong generalization across spatial scales. 
Compared to CNN baselines, BNO delivers more robust predictive performance, particularly in regions characterized by complex flow dynamics. 
Its operator-theoretic design, leveraging a Koopman-based formulation in Bnach spaces, fosters global coherence and mitigates overfitting, a common limitation of CNNs. 
Furthermore, the ability to successfully transfer BNO-trained parameters to the CNN architecture---while retaining high-resolution accuracy---underscores the robustness and reusability of the BNO's learned representations. 
In contrast, CNN-trained parameters fail to transfer effectively to the BNO framework, reflecting a more limited generalization capacity. 
These findings position BNO as a powerful framework for flow field reconstruction in data-scarce environments, especially when high-resolution labels are unavailable, and underscore its potential for deployment in real-world applications requiring scale-agnostic learning. 

While BNO architectures configured with multiple Banach layers have the potential to capture complex spatiotemporal dynamics, they suffer from significant numerical and stability challenges that hinder their predictive reliability.
Specifically, the failure of Cholesky decomosition---due to the emergence of non-SPD matrices---highlights the compounded effects of numerical instability, rank deficiency, ill-conditioning, and gradient issues as network depth increases. 
Empirical evaluations further indicate that a single Banach layer not only circumvents these issues but also consistently outperforms deeper configurations in zero-shot super-resolution tasks, particularly in critical flow regions such as shear layers and wakes. 
These findings suggest that increasing model depth in BNOs should be approached with caution, and future research should focus on enhancing stability through improved regularization, alternative decomposition techniques, or architectural innovations that preserve the SPD structure throughout training. 

\section*{Acknowledgements}
The author thanks the anonymous reviewers for their comments and suggestions that helped to promote the quality and clarity of the manuscript.  
This research was supported by the Division of Research and Innovation Partnership Commitments (RIPS) at Northern Illinois University. 


\section*{Data availability}
The data that support the findings of the present work are available from the corresponding author upon reasonable request. 

\section*{References}
\bibliographystyle{elsarticle-num}
\bibliography{refs}

\end{document}